# Geschlechtsübergreifende Maskulina im Sprachgebrauch
## Eine korpusbasierte Untersuchung zu lexemspezifischen Unterschieden

*Carolin Müller-Spitzer, Samira Ochs, Jan Oliver Rüdiger, Sascha Wolfer*

This study examines the distribution and linguistic characteristics of generic masculines (GM) in contemporary German press texts. The use of masculine personal nouns to refer to mixed-gender groups or unspecified individuals has been widely debated in academia and the public, with conflicting perspectives on its gender-neutrality. While psycholinguistic studies suggest that GM is more readily associated with male referents, corpus-based analyses of its actual use remain scarce. We investigate GM in a large corpus of press texts, focusing on lexeme-specific differences across different types of personal nouns. We conducted manual annotations of the whole inflectional paradigm of 21 personal nouns, resulting in 6,195 annotated tokens. Our findings reveal considerable differences between lexical items, especially between passive role nouns and prestige-related personal nouns. On a grammatical level, we find that GM occurs predominantly in the plural and in indefinite noun phrases. Furthermore, our data shows that GM is not primarily used to denote entire classes of people, as has been previously claimed. By providing an empirical insight into the use of GM in authentic written language, we contribute to a more nuanced understanding of its forms and manifestations. These findings provide a solid basis for aligning linguistic stimuli in psycholinguistic studies more closely with real-world language use.

## Inhalt

# 1   Einleitung und Motivation

Dreh- und Angelpunkt der Debatte um genderinklusive Sprache im Deutschen sowie in anderen Sprachen mit Genusmarkierung bei Nomen („grammatical gender languages" vgl. Corbett 2013) ist das sogenannte generische bzw. geschlechtsübergreifende Maskulinum (= GM)[1]. Der Begriff beschreibt die Verwendung maskuliner Personenbezeichnungen, um nicht nur spezifisch auf männliche Personen zu verweisen, sondern auch geschlechtsübergreifend auf Einzelpersonen oder Gruppen – insbesondere dann, wenn deren Geschlechtsidentität unbekannt oder irrelevant ist (s. z.B. Diewald 2018; Kotthoff/Nübling 2024a: 91–127). Die Verwendung maskuliner Personenbezeichnungen zur geschlechtsübergreifenden Referenz ist schon lange Gegenstand kontroverser wissenschaftlicher und öffentlicher Debatten (in Auswahl: Müller-Spitzer 2022a, 2022b; Pusch 1984; Simon 2022; Trutkowski/Weiß 2023; Meineke 2023). Einige Forscher*innen sehen diese Verwendung nicht als geschlechtsneutrale Referenz an (z.B. Acke 2019: 308; Diewald 2018; Hellinger/Bußmann 2003: 160–161). Andere halten dagegen die geschlechtsübergreifende Verwendung von Maskulina per se für genderneutral (Eisenberg 2020; z.B. Meineke 2023; Zifonun 2018).

Diese unterschiedlichen Haltungen oder Deutungen werden auch in der begrifflichen Fassung des Phänomens reflektiert: Während Forscher*innen, die GM als geschlechtsneutrale Referenz ansehen, eher die Termini *generisches Maskulinum* (z.B. Eisenberg 2022; Glück 2020) oder *genderneutrales Maskulinum* (z.B. Meineke 2023) verwenden, wird in der genderlinguistischen Forschung, die die Genderneutralität dieser Art der Referenz kritisch diskutiert, meist für den Begriff *sogenanntes generisches Maskulinum* (z.B. Diewald 2018) oder *geschlechtsübergreifendes Maskulinum* (z.B. Kotthoff/Nübling 2024a: 104) plädiert.

Die Gründe für die Position, dass GM grundsätzlich genderneutral sind, beruhen z.B. auf dem Konzept der konversationellen Implikatur (Becker 2008; s. zu einer anderen Interpretation Diewald 2018, 2025a, 2025b), dem der ‚Unmarkiertheit' maskuliner Personenbezeichnungen (Eisenberg 2020; vgl. im Kontrast dazu Haspelmath 2006) oder auf der Annahme historischer Kontinuitäten in der Verwendung (Trutkowski/Weiß 2023). Viele psycholinguistische und kognitionswissenschaftliche Studien kommen jedoch zu dem Ergebnis, dass GM gedanklich schneller und leichter mit männlichen als mit weiblichen oder non-binären Personen assoziiert werden können, d.h. einen *male bias* aufweisen (z.B. Glim et al. 2023; Gygax et al. 2008; Körner et al. 2022; Zacharski/Ferstl 2023; für einen umfassenden Überblick s. Körner 2025). Dieser *male bias* scheint zumindest teilweise auf die grammatikalischen Eigenschaften des Deutschen zurückzuführen zu sein, in denen die maskuline Form diese Doppelfunktion – zum einen die geschlechtsspezifisch und zum anderen die geschlechtsübergreifend intendierte Referenz – erfüllt (Garnham et al. 2012). Zudem legen Gygax et al. (2016) nahe, dass Geschlechterstereotype und die realen Geschlechterverteilungen in den jeweiligen gesellschaftlichen Gruppen, auf die mit den einzelnen Personenbezeichnungen referiert wird, diese Effekte moderieren.

---

[1] Im Folgenden kürzen wir den Terminus als *GM* ab, auch wenn er sich auf den Plural *generische Maskulina* bezieht.

Ein häufiges Argument gegen die Aussagekraft psycholinguistischer Studien zum GM ist, dass aus ihnen keine generellen Aussagen abzuleiten sind, da nur eine bestimmte Art von GM in ihnen getestet würde (s. Körner 2025 für eine ausführliche Diskussion und Widerlegung der diversen Kritikpunkte an psycholinguistischen Untersuchungen). So werden bei Gygax et al. (2008) beispielsweise nur GM in definiten Nominalphrasen im Plural getestet (z.B. *Die Spione kamen aus dem Besprechungsraum*). „Tests dieser Art", so Zifonun (2018: 51) sagen allerdings „nichts […] über eine generell mit dem generischen Maskulinum assoziierte mentale Sexus-Zuweisung" aus.[2] Der indefinit-unspezifische bzw. „essentielle" Gebrauch der Art *Die Situation von Ausländern in Deutschland ist sehr schwierig* (Zifonun 2018: 51, Fußnote 8) werde damit nicht untersucht, weshalb keine generellen Aussagen zum GM möglich seien. Der Vorteil beim „Gebrauch von Oberbegriffen", so vermutet Zifonun, sei aber wahrscheinlich „der Hauptgrund, warum viele sich des generischen Maskulinums bedienen" (Zifonun 2018: 50). Sie betont:

> „Diese Irrelevanz von Gender- oder Sexusmerkmalen wird von den Gegnern des Genderns generell zur Verteidigung des generischen Maskulinums in Anspruch genommen. Referenzsemantisch spricht einiges für diese Position, zumindest soweit es um den essentiellen oder indefinit-unspezifischen Gebrauch einschlägiger maskuliner Nominalphrasen geht." (Zifonun 2018: 51)

Daher stellt sich die Frage, in welcher Art von Nominalphrasen GM tendenziell eingebettet sind. Gerade die nicht abreißende Kritik an psycholinguistischen Studien, die definite Nominalphrasen testen, macht eine empirische Überprüfung dieses Aspekts relevant.

Es gibt in der Literatur verschieden gelagerte Postulate über ‚typische' geschlechtsübergreifende Maskulina im Sprachgebrauch, von denen wir hier zwei exemplarisch zitieren wollen. Meineke (2023) konstatiert:

> Eine Hauptdomäne des genderneutralen Maskulinums waren und sind generelle Aussagen, in denen eine Gattung ungeachtet des Geschlechts der zugehörigen Individuen charakterisiert wird: *Der Schwabe gilt als fleißig. (Die) Berliner sind bekanntlich nicht auf den Mund gefallen.* (Meineke 2023: 55. Hervorh. im Original)

Wegener wiederum meint, dass GM „fast nur im Plural" vorkommen, was „semantische und formale Gründe" habe: „Nur Gruppen können männlich *und* weiblich, gemischt also sein [sic], ein Nomen im Singular kann nur als männlich *oder* weiblich vorgestellt werden." (Wegener 2024: 39, Hervorh. im

---

Orig.)[3] Insgesamt sind solche Aussagen wie die über das Auftreten von GM – ob sie vorwiegend definit oder indefinit, in generellen Aussagen oder nicht, im Singular oder Plural verwendet werden – zunächst Annahmen.

Mit der vorliegenden Studie wollen wir die Forschungsdiskussion um das GM mit empirisch fundierten Korpusanalysen anreichern und die o.g. Annahmen überprüfen. Diese sind noch einmal zusammengefasst: (1) GM kommen vor allem in indefinit-unspezifischen Nominalphrasen vor; (2) GM kommen im Sprachgebrauch fast nur im Plural vor; (3) GM werden vor allem als Gattungsbezeichnungen verwendet. Unsere Studie sehen wir als Beitrag in einer Reihe korpuslinguistischer Arbeiten (Link 2024; Müller-Spitzer et al. 2024, 2025; Müller-Spitzer/Ochs 2023; Ochs/Rüdiger 2025; Schmitz 2024; Schmitz et al. 2023; Waldendorf 2023), die versuchen, von *Postulaten über den Gebrauch* von Personenbezeichnungen zu *empirisch fundiertem Wissen zum Gebrauch* zu gelangen. Im Unterschied zu Müller-Spitzer et al. (2024) setzen wir in dieser Untersuchung auf der Lexemebene an. Das bedeutet, dass wir nicht alle vorkommenden GM in nach bestimmten Kriterien ausgewählten Texten untersuchen, sondern von konkreten Einzellexemen ausgehen (vgl. für eine lexembasierte Analyse von Genderzeichen Ochs/Rüdiger 2025). Dies ist für die vorliegende Studie besonders wichtig, da wir die oben zitierten Annahmen zu ‚typischen' GM hinsichtlich Numerus oder (In-)Definitheit empirisch überprüfen wollen. Dabei wissen wir aber, dass z.B. der Numerusgebrauch nicht zufällig unter Lexemen verteilt ist. Manche Lexeme kommen z.B. vorwiegend im Plural, manche vorwiegend im Singular vor (Reifegerste et al. 2017: 477), sodass Aussagen zum Numerus geschlechtsübergreifender Maskulina nur lexemspezifisch sinnvoll sind. Zum anderen ist bekannt, dass Lexeme unterschiedlich ‚genderisiert' sind:

> Hinzu kommt, dass Lexeme ein sog. soziales Geschlecht haben können (genderisiert sein können), das sich aus außersprachlichen Geschlechterverteilungen oder -vorstellungen (die oft historisch befrachtet sind) speist: *Piloten* und *Professoren* werden eher männlich gelesen als *Touristen*, *Lehrer* und *Patienten* oder gar *Erzieher*, *Kosmetiker* und *Altenpfleger*." (Kotthoff/Nübling 2024a: 108)

Wenn wir also die Häufigkeit geschlechtsübergreifender Maskulina mit der Häufigkeit geschlechtsspezifischer Maskulina und Feminina vergleichen wollen, ist eine lexembasierte Kontrolle methodisch sinnvoll, da unterschiedliche (Arten von) Personenbezeichnungen unterschiedliche Verteilungen aufweisen könnten. Mit dieser Herangehensweise lässt sich auch überprüfen, wie bzw. ob sich ausgehend von einzelnen Lexemen generalisierbare Aussagen zur Häufigkeit von GM treffen lassen. Krome (2021: 29, Fußnote 7) nennt beispielsweise das Lexem BÜRGER[4] als ‚paradigmatischen Fall' und stützt ihre Einordnungen zur Häufigkeit des GM im Sprachgebrauch ausschließlich auf eine stichprobenartige Analyse der Belege zu diesem Wort.[5] Die recht allgemeingültig formulierten Schlussfolgerungen aus der Analyse, dass nach wie vor das GM am „weitaus meisten verbreitet" sei, und dass „Ansätze zu geschlechtergerechter Schreibung" in Zeitungstexten „weniger präsent" seien, basieren demnach auf einer einzelnen Fallanalyse. Für uns schließt sich hier die wichtige Forschungsfrage an, ob geschlechtsübergreifende und -spezifische Maskulina tatsächlich so gleichmäßig über verschiedene Lexeme verteilt sind, dass die Untersuchung eines einzelnen Items ausreicht, um solch weitreichende Schlussfolgerungen zu ziehen.

---

[3] Abgesehen davon, dass hier innerhalb eines binären Geschlechterverständnisses argumentiert wird, liegt auch eine terminologisch fragwürdige Vermischung grammatischer und außersprachlicher Ebenen vor.
[4] Im Folgenden werden für Lexeme immer dann Kapitälchen verwendet, wenn die Grundform stellvertretend für alle Formen des Paradigmas genannt wird (wie BÜRGER für *Bürger, Bürgers, Bürgern, Bürgerin, Bürgerinnen*).

Unser Beitrag ist folgendermaßen gegliedert: In Abschnitt 2 erläutern wir unsere Forschungsfragen und Hypothesen. Die Datenerhebung und die Methode werden in Abschnitt 3 vorgestellt. Wir beschreiben dort auch das Annotationsschema und damit die semantisch-pragmatische Operationalisierung zentraler Konzepte wie geschlechtsübergreifendes bzw. -spezifisches Maskulinum. Die Ergebnisse folgen in Abschnitt 4, zunächst zu den Anteilen von GM nach Lexemen (4.1), zu Numerus und Definitheit (4.2) und schließlich zur Einordnung als Gattungsbezeichnungen (4.3). Der Beitrag endet in Abschnitt 5 mit Diskussion und Ausblick, in dem insbesondere mögliche Anschlussfragen für die psycholinguistische Forschung skizziert werden.

## 2   Forschungsfragen und Hypothesen

Unsere Studie untersucht, ob die lexikalisch hochdiversifizierten Personenbezeichnungen unterschiedliche Verwendungsmuster zeigen. Im Fokus steht dabei die Frage, wie häufig geschlechtsübergreifende Maskulina im Vergleich zu geschlechtsspezifischen Maskulina und Feminina auftreten. Gleichzeitig analysieren wir, welche der eingangs zitierten Annahmen zum GM im Sprachgebrauch bestätigt werden können. Unsere Forschungsfragen lauten:

- Zeigen sich bei der Verwendung von GM sowie geschlechtsspezifischen Maskulina und Feminina lexemspezifische Unterschiede? Lassen sich dabei Muster erkennen (z.B. hinsichtlich Typen von Personenbezeichnungen)?
- In welchem Numerus kommen GM vor?
- Wie häufig sind GM in definite und indefinite Nominalphrasen (= NP) eingebettet?
- Wie häufig kommen GM als Gattungsbezeichnungen vor (nach Meineke 2023)?

Um Muster in der Verwendung von Personenbezeichnungen zu identifizieren, ist es sinnvoll, Gruppen von Lexemen zu bilden. Wie sich Personenbezeichnungen allerdings klassifizieren lassen, v.a. um (korpus-)analytisch damit zu arbeiten, ist bisher wenig diskutiert worden. Eine in der Genderlinguistik verbreitete Unterscheidung ist die von Berufen und Rollen. Beispielsweise schreiben Kotthoff und

---

[5] Bestätigt durch persönliche Kommunikation mit der Autorin.

Nübling, dass Berufsbezeichnungen stärker männlich genderisiert seien als sog. Rollenbezeichnungen (Kotthoff/Nübling 2024a: 138). Im Detail ist diese Unterscheidung allerdings nicht immer so einfach, wie sie zunächst erscheint, und es existieren etliche Graubereiche. So kann z.B. eine *Studentin* auch die Rolle einer *Lehrerin* haben (im Tutorium oder bei der Nachhilfe), *Präsident* kann ein prestigeträchtiger Beruf sein (wie im Falle des Bundespräsidenten), oder eine ehrenamtlich ausgeführte Rolle im Sportverein. Besonders in Bereichen wie Sport, Kunst und Kultur kann häufig nur anhand des Kontextes entschieden werden, ob es sich um eine bezahlte Tätigkeit (z.B. bezahlter Fußballspieler) oder eine Freizeitaktivität handelt (z.B. Hobby-Fußballspieler im Sportverein; vgl. Bröder/Rosar 2025). Auch der Kontext wirkt auf das Ausmaß der Genderisierung (Bülow/Jakob 2017; Rothmund/Scheele 2004): Die gleiche Personenbezeichnung (SPORTLER) wirkt beispielsweise in einem Eishockey-Kontext männlicher genderisiert als in einem Yogakontext (Braun et al. 1998).

Ein Klassifizierungsversuch, der über die Berufe und Rollen hinausgeht, findet sich bei Bühlmann (2002: 174). Sie stellt insgesamt vier Bezeichnungskategorien auf: Prestige, Aktivitäts-, Passivitäts- und Bevölkerungsbezeichnungen. Im Folgenden erläutern wir kurz die einzelnen Kategorien nach Bühlmann, ergänzt um eigene Erweiterungen der Definitionen, wie wir sie für die Gruppeneinteilung von Personenbezeichnungen in unserer Studie angewendet haben:

- Als **Prestigebezeichnungen** werden solche Personen- oder Berufsbezeichnungen eingeordnet, die ein gewisses Prestige besitzen (v.a. Politik, hochrangige Führungspositionen), z.B. KANZLER, MINISTER.
- **Aktivbezeichnungen** sind Bezeichnungen, die ein gewisses Maß an Ausbildung bzw. Aktivität voraussetzen. Meistens sind dies Berufe, aber auch bestimmte Hobbys oder Freizeitaktivitäten, z.B. LEHRER, MALER.
- Als **Passivbezeichnungen** gelten Bezeichnungen, die nichts oder nicht viel voraussetzen; v.a. Rollen, in denen sich Personen mehr oder weniger unfreiwillig oder zufällig befinden. In einem bestimmten Kontext kann jede*r diese Rolle erfüllen, ohne Ausbildung oder besondere Aktivität, z.B. BÜRGER, SCHÜLER.
- **Bevölkerungsbezeichnungen** sind Bezeichnungen für Bevölkerungsgruppen, z.B. geographisch oder politisch, z.B. ENGLÄNDER, KARLSRUHER.

Diese Kategorisierung dient uns als Grundlage zur Auswahl und Analyse einer bestimmten Menge an Personenbezeichnungen (vgl. Kapitel 3.2). Dass die „Einteilung […] nicht immer eindeutig und deshalb auch nicht hundertprozentig exakt [ist]" (Bühlmann 2002: 174), ist bei einer solchen qualitativ-semantischen Gruppierung unvermeidbar. Dennoch hilft es dabei, Schlüsse über kategorienabhängige Verwendungsweisen zu ziehen (vgl. Kapitel 4.1).

Unsere zu überprüfenden Hypothesen lauten demnach:

**H1:** Die Anteile geschlechtsübergreifender und geschlechtsspezifischer Maskulina bzw. Feminina (Movierungen) variieren je nach Lexem.

**H2:** Die Bezeichnungskategorien unterscheiden sich hinsichtlich der Verteilung geschlechtsübergreifender und -spezifischer Maskulina/Feminina.

Prestigeträchtige Berufe oder Positionen mit einem hohen Männeranteil gelten als besonders stark männlich genderisiert. Allgemeine Rollenbezeichnungen wie Einwohner, Bürger oder Teilnehmer hingegen werden eher als neutral wahrgenommen und rufen tendenziell gemischtgeschlechtliche Vorstellungen hervor, vermutlich auch, da die realen Bevölkerungsanteile in diesen Gruppen ausgewogen sind (Kotthoff/Nübling 2024a: 136, 138, s. dort auch eine Zusammenfassung der entsprechenden Studien). Dementsprechend lauten unsere Subhypothesen zu H2:

> **H2.1:** Personenbezeichnungen der Kategorien *Aktiv-* oder *Prestigebezeichnung* werden überwiegend als geschlechtsspezifische Maskulina verwendet.

> **H2.2:** Personenbezeichnungen der Kategorien *Bevölkerungs-* und *Passivbezeichnung* werden überwiegend als geschlechtsübergreifende Maskulina verwendet.

Die nächsten beiden Hypothesen untersuchen Faktoren, die über die Bezeichnungskategorie hinaus das Auftreten der Referenzialitätstypen (geschlechtsübergreifend vs. -spezifisch) beeinflussen könnten, nämlich Numerus und Definitheit (vergl. hierzu Kapitel 1). Die Hypothesen H3 und H4 ergänzen damit die in H1 untersuchte lexikalische Dimension durch grammatische Variablen:

**H3:** Numerus und Referenzialitätstyp stehen in einem signifikanten Zusammenhang.

> **H3.1:** Geschlechtsübergreifende Maskulina treten überwiegend im Plural auf.

> **H3.2:** Geschlechtsspezifische Referenzen treten überwiegend im Singular auf.

**H4:** Definitheit und Referenzialitätstyp stehen in einem signifikanten Zusammenhang.

> **H4.1:** Geschlechtsübergreifende Maskulina treten überwiegend in indefiniter NP auf.

> **H4.2:** Geschlechtsspezifische Referenzen treten überwiegend in definiter NP auf.

Um außerdem noch die Hypothese zu überprüfen, dass GM vorwiegend als Gattungsbezeichnungen vorkommen, wurden alle als geschlechtsübergreifende Maskulina annotierten Verwendungen dahingehend kategorisiert, ob sie im Sinne von Meineke (2023: 55) als Gattungsbezeichnungen fungieren. Die entsprechende Hypothese lautet:

**H5:** Geschlechtsübergreifende Maskulina treten vorwiegend als Gattungsbezeichnungen auf.

Insgesamt erlauben diese Analysen damit differenzierte, empirisch fundierte Aussagen über häufige bzw. ‚typische' Verwendungsweisen des GM. Die statistische Überprüfung von H1–H4 erfolgt in den Kapiteln 4.1 und 4.2. Die Detailanalysen zu geschlechtsübergreifenden Maskulina als Gattungsbezeichnungen (H5) werden in Kapitel 4.3 vorgestellt.

# 3  Datenerhebung und Methode

## 3.1  Korpusgrundlage

Als Korpusgrundlage verwenden wir Pressetexte aus unterschiedlichen Quellen, die im Deutschen Referenzkorpus (DeReKo; Kupietz et al. 2010, 2018) zur Verfügung stehen (zur Einordnung von Pressetexten als Quellen des Gebrauchsstandards vgl. Eisenberg 2007: 217; Klosa 2011: 14; Storjohann 2005: 63): Einerseits verwenden wir Texte der Deutschen Presseagentur (dpa), da diese besonders weite Verbreitung und Rezeption finden und politisch neutral orientiert sind. Andererseits analysieren wir Zeitschriftenquellen, die von unterschiedlichen Verlagen herausgegeben werden und sich an verschiedene Zielgruppen richten: *Brigitte*, *Psychologie Heute* und *Zeit Wissen* (s. zu einer vergleichbaren Vorgehensweise Müller-Spitzer et al. 2024). Wir analysieren ein zufälliges Textsample aus den Jahrgängen 2006 bis 2020, das die anvisierten Personenbezeichnungen enthält. Es wurden ganze Texte annotiert, da für die Interpretation des Referenztyps – also ob eine maskuline Personenbezeichnung geschlechtsübergreifend oder -spezifisch verwendet wird – möglichst viel Kontext erforderlich ist (Müller-Spitzer et al. 2024). Durch diese Vorgehensweise, insbesondere die Zufallsauswahl aus einer großen Textmenge, können wir generalisierbare Aussagen zur Verwendung geschlechtsübergreifender Maskulina in dieser Art von Pressetexten treffen. Gleichzeitig ist uns bewusst, dass unsere Untersuchung nur einen von vielen möglichen Beiträgen zum umfassenden Verständnis der Verwendung von GM im Sprachgebrauch leisten kann und weitere Untersuchungen an anderen Textsorten und -formaten notwendig sind.

## 3.2  Auswahl und Annotation der Items

Die Auswahl der Personenbezeichnungen, die im Textsample annotiert wurden, erfolgte anhand einer Liste, die uns von der Duden-Redaktion zur Verfügung gestellt wurde. Diese enthält ca. 10.000 movierbare Personenbezeichnungen (der Form *Bürger – Bürgerin*, *Kollege – Kollegin)*, die wir zunächst doppelt danach annotiert haben, welcher Bezeichnungskategorie sie nach Bühlmann (2002: 174) angehören. Zusätzlich wurden die jeweiligen Frequenzen für die maskulinen und femininen Formen in DeReKo 2024-I ermittelt.[6] Aus den Personenbezeichnungen wurden dann diejenigen ausgewählt, die sowohl als Feminina als auch als Maskulina zu den häufigsten zehn Bezeichnungen einer Kategorie gehörten.

---

[6] Eine solche Frequenzermittlung sollte vorzugsweise durch die Addition der Tokenfrequenzen der einzelnen Formen eines Flexionsparadigmas erfolgen, anstatt auf der Lemmatisierung zu basieren. Der Grund dafür ist, dass gängige Lemmatisierer movierte Formen fälschlicherweise dem maskulinen Paradigma zuordnen könnten.

Dass wir nur solche Personenbezeichnungen in unserem Auswahlprozess einbezogen haben, zu denen es sowohl eine feminine wie maskuline Form gibt, liegt an der zentralen Untersuchungskategorie *geschlechtsübergreifendes Maskulinum*. Generell wenden wir die Kategorisierung GM nur auf die geschlechtsübergreifende Verwendung maskuliner Personenbezeichnungen an, zu denen es eine Movierung gibt (*Lehrer-Lehrerin, Bürger-Bürgerin*), oder eine andere Paarform vorliegt (*Kranker-Kranke, Jugendlicher-Jugendliche*. Letztere Typen von Personenbezeichnungen sind im Plural genusneutral und können in dieser Form als geschlechtsabstrahierend gelten. Genauso sehen wir anders als z.B. Kopf (2023: 197) maskuline Epikoina wie *Mensch* oder *Fan* nicht als GM an, sondern als geschlechtsneutrale Referenzform (genauso wie Epikoina in anderen Genera). Für diese Studie wurden daher nur solche Lexeme ausgewählt, die sich sowohl im Singular wie im Plural im Maskulinum und Femininum unterscheiden.

Folgende 21 Items sind Teil unserer Untersuchung[7]:

- 6 Prestigebezeichnungen: BÜRGERMEISTER, CHEF, MINISTER, PFARRER, PRÄSIDENT, RICHTER
- 6 Aktivbezeichnungen: GESCHÄFTSFÜHRER, KÜNSTLER, LEHRER, LEITER, MITARBEITER, SPIELER
- 5 Passivbezeichnungen: BÜRGER, FREUND, GASTGEBER, SCHÜLER, TEILNEHMER
- 4 Bevölkerungsbezeichnungen: BERLINER, HAMBURGER, SCHWEIZER, WIENER

Insgesamt wurden pro Item mindestens 100 Tokens aus mindestens 25 Texten zufällig gesampelt, damit eine Streuung über die Texte gewährleistet ist.[8] Falls also beispielsweise nach zehn Texten bereits 100 Vorkommen eines Lexems erreicht waren, wurde weitergesampelt, bis 25 Texte in unser Korpus eingeflossen sind. Ebenso wurde weiter gesampelt, falls in 25 Texten noch keine 100 Vorkommen der anvisierten Personenbezeichnung erreicht waren. In einem weiteren Schritt wurden die zufällig ausgewählten Texte für die Annotator*innen vorgesichtet, um *false positives* auszuschließen (bspw. Texte, in denen Richter oder Freund als Eigennamen vorkamen). Tabelle 1 zeigt eine Übersicht über die Gesamtzahl der so gesampelten Texte und die übereinstimmend annotierten Tokens pro Personenbezeichnung. Die Spalte *aufgenommene Texte* zeigt die Menge an Texten, die im Sample verblieben, die Spalte *abgelehnte Texte* zeigt, wie viele Texte ausgeschlossen wurden. Besonders viele Texte mussten bei den Bevölkerungsbezeichnungen abgelehnt werden, da diese im überwiegenden Teil der Fälle adjektivisch gebraucht werden (z.B. *Schweizer Nationalrat, Wiener Schnitzel*). Sie sind damit, neben Gastgeber, auch die einzigen Items im Datensatz, die nicht die Mindestanzahl von 100 annotierten Tokens erreichen (vgl. Tabelle 1, letzte Spalte, gezeigt ist allerdings nur die Anzahl übereinstimmend annotierter Tokens). Da der Studie ein ausführlicher Designplan zugrunde lag, an der wir uns strikt gehalten haben, und der Sampling-Prozess außerdem sehr zeit- und ressourcenintensiv ist, wurde kein Nachsampling für die somit unterrepräsentieren Bevölkerungsbezeichnungen vorgenommen. Stattdessen werden sie in den nachfolgenden Analysen mit größerer Vorsicht interpretiert als die übrigen Items.

---

[7] Die 21 Lexeme ergeben sich aus der Schnittmenge der jeweils zehn häufigsten Maskulina und Feminina pro Kategorie in DeReKo. Wir haben für jede Kategorie die Top10-Maskulina und die Top10-Feminina verglichen und jeweils nur diejenigen aufgenommen, die in beiden Listen vorkommen. Je nach Kategorie überlappen sich unterschiedlich viele Paare, daher kommen 6 Prestige-, 6 Aktiv-, 5 Passiv- und 4 Bevölkerungsbezeichnungen zusammen.

[8] Dies wäre bei mündlichen Korpora zum Deutschen nicht möglich gewesen, auch wenn diese Korpora in den letzten Jahren deutlich größer geworden sind. Beispielsweise sind im *Forschung- und Lehrkorpus Gesprochenes Deutsch* (FOLK; vgl. Reineke et al. 2023) in der Version 2.22 (Juni 2024) die Lexeme in der maskulinen Form als Lemma folgendermaßen belegt. Prestigebezeichnungen: *Bürgermeister (100), Chef (177), Minister (25), Pfarrer (20), Präsident (75), Richter (26)*; Aktivbezeichnungen: *Geschäftsführer (20), Künstler (29), Lehrer (245), Leiter (86*; alle Vorkommen inkl. *Leiter zum Springturm* usw.*), Mitarbeiter (206), Spieler (231)*; Passivbezeichnungen: *Bürger (202), Freund (693), Gastgeber (8), Schüler (328), Teilnehmer (17)*; Bevölkerungsbezeichnungen (alle Vorkommen inkl. adjektivischer Verwendung): *Berliner (51), Hamburger (80), Schweizer (55), Wiener (18)*. Eine Zufallsauswahl mit mindestens 100 Vorkommen aus mindestens 25 Quellen wäre hier also nicht möglich.

| Personenbezeich-nung | aufgenommene Texte | abgelehnte Texte | Gesamtzahl Texte | Anzahl übereinstimmend annotierter Tokens |
|---|---|---|---|---|
| Prestigebezeichnungen | | | | |
| Bürgermeister | 102 | 0 | 102 | 170 |
| Chef | 137 | 3 | 140 | 444 |
| Minister | 94 | 1 | 95 | 180 |
| Pfarrer | 104 | 1 | 105 | 115 |
| Präsident | 140 | 0 | 140 | 358 |
| Richter | 80 | 60 | 140 | 157 |
| Aktivbezeichnungen | | | | |
| Geschäftsführer | 124 | 0 | 124 | 174 |
| Künstler | 140 | 0 | 140 | 264 |
| Lehrer | 140 | 0 | 140 | 576 |
| Leiter | 133 | 7 | 140 | 242 |
| Mitarbeiter | 136 | 4 | 140 | 611 |
| Spieler | 129 | 5 | 134 | 251 |
| Passivbezeichnungen | | | | |
| Bürger | 137 | 3 | 140 | 301 |
| Freund | 136 | 4 | 140 | 1139 |
| Gastgeber | 91 | 0 | 91 | 86 |
| Schüler | 133 | 0 | 133 | 573 |

| TEILNEHMER | 140 | 0 | 140 | 408 |
| --- | --- | --- | --- | --- |
| Bevölkerungsbezeichnungen | | | | |
| BERLINER | 29 | 111 | 140 | 56 |
| HAMBURGER | 14 | 126 | 140 | 33 |
| SCHWEIZER | 40 | 100 | 140 | 46 |
| WIENER | 7 | 128 | 135 | 11 |
| *gesamt* | *2.186* | *553* | *2.739* | *6.195* |

*Tabelle 1: Anzahl akzeptierter und abgelehnter Texte. Sortierung nach Bezeichnungskategorie und dann alphabetisch nach Personenbezeichnung.*

Die verbliebenen Texte und Tokens wurden dann in der eigens programmierten Software *QuickAnnotator* (Rüdiger 2025) von zwei studentischen Annotator*innen annotiert, die zunächst in Testrunden trainiert wurden. Die Zielitems waren im *QuickAnnotator* hervorgehoben, sodass die Annotator*innen direkt erkennen konnten, welche Bezeichnungen zu annotieren sind. Da dadurch die Entscheidung obsolet wurde, ob es sich bei einem Item überhaupt um eine Personenbezeichnung handelt oder nicht (vgl. die Unsicherheit von 16% bei der Entscheidung, ob eine Personenreferenz vorliegt oder nicht, bei Müller-Spitzer et al. 2024: 5), ist das Inter-Annotator-Agreement mit 91,5% sehr hoch. Abbildung 1 zeigt, welche Kategorien in Abhängigkeit voneinander annotiert werden konnten.

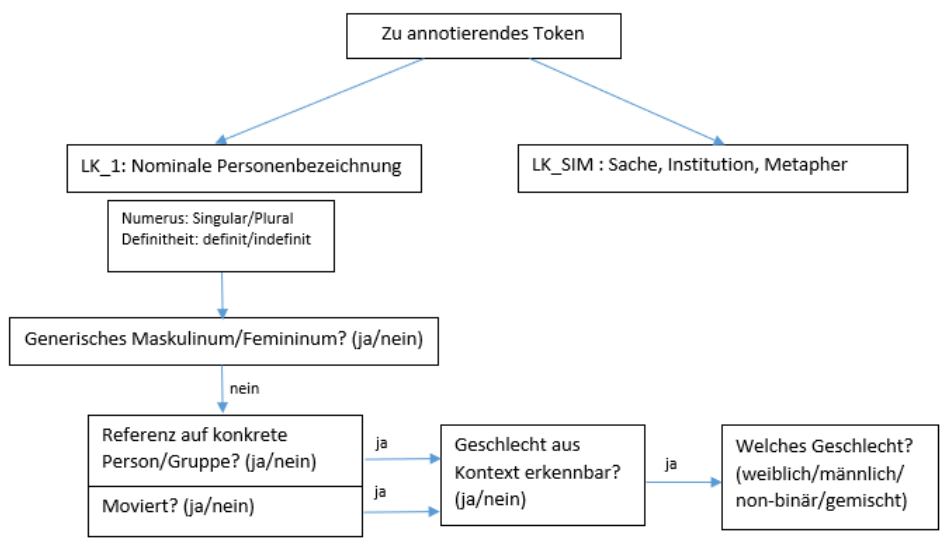

*Abbildung 1: Entscheidungsbaum für die Annotation der Personenbezeichnungen.*

Zunächst musste bei jeder hervorgehobenen Personenbezeichnung eingeordnet werden, ob sie sich auf eine natürliche Person bezieht oder für eine Sache oder Institution steht bzw. metaphorisch gebraucht wird – etwa in Formulierungen wie **Gastgeber** Frankreich sind [die Beutel] gewidmet oder

*Wenn wir [den Atem] zu unserem **Freund** machen*. Fälle dieser Art wurden als Sache, Institution bzw. Metapher (LK_SIM) ausgezeichnet und nicht weiter annotiert.

Weitere Ausdifferenzierungen zur Verwendung wurden nur für Bezeichnungen natürlicher Personen getroffen. Diese wurden entsprechend der oben aufgeführten Hypothesen zunächst nach **Numerus** (Singular/Plural) und **Definitheit** (indefinit/definit) eingeordnet. Definitheit im referenzsemantischen Sinne ist insgesamt ein vielschichtiges linguistisches Konzept, das je nach Perspektive unterschiedlich definiert und interpretiert werden kann.[9] Für eine empirische Studie wie die vorliegende ist es essentiell, eine nachvollziehbare Operationalisierung komplexer Konzepte zu entwickeln. Diese sollte einerseits nicht zu anspruchsvoll sein, um studentische Annotator*innen in angemessener Zeit sicher mit dem Annotationsschema vertraut zu machen und eine hohe Übereinstimmung bei der Annotation zu gewährleisten. Andererseits muss sie feine und granulare Unterscheidungen ermöglichen, die klar voneinander abtrennbar sind. Dies ist nicht nur für den Annotationsprozess selbst entscheidend, sondern auch für die Auswertung und Reproduzierbarkeit der Ergebnisse. Daher haben wir eine bewusst vereinfachte Operationalisierung gewählt, die sich an Artikeln und Quantifikatoren orientiert (Pettersson 2011: 68; Pustka 2013). Definitheit umfasst in unserer Annotation Nominalphrasen mit bestimmtem Artikel, aber auch mit Demonstrativ- oder Possessivartikel (anders als z.B. Pafel 2020: 31–46, der Demonstrativa gesondert behandelt) sowie absolute Quantifikatoren (d.h. Zahlwörter). Indefinitheit hingegen schließt Nominalphrasen mit unbestimmtem Artikel, Nullartikel sowie relativen Quantifikatoren (z.B. *einige, manche, alle*) ein.[10]

Die zentrale Annotationskategorie in unserer Studie ist die Einordnung einer Personenbezeichnung als **geschlechtsübergreifende** oder -**spezifische Referenz**. Eine manuelle Annotation ist hierbei unerlässlich, da nur so geschlechtsübergreifende und -spezifische Referenzen verlässlich voneinander unterschieden werden können (vgl. Müller-Spitzer et al. 2024; Sökefeld et al. 2023). Obwohl das GM ein zentrales Konzept der Genderlinguistik und der Debatte über genderinklusive Sprache ist, fehlt es häufig an klaren Operationalisierungen. Wir stützen uns hier auf die Einordnung von Kotthoff und Nübling (2024b: 104):

> „*Generisch* wird somit in der Bedeutung von *geschlechtsübergreifend* oder -*inklusiv*, *geschlechtsneutral*, -*indifferent* oder -*abstrahierend* gefasst und bildet die Opposition zu

---

*geschlechtsspezifisch* oder *geschlechtdefinit*." (vgl. auch Pettersson 2011; Zifonun 2018: 50; sowie Bröder/Rosar 2025, die sich bei GM allerdings auf generische Referenzen im referenzsemantischen Sinne beschränken)

Bei der Annotation muss also aus dem Textzusammenhang interpretiert werden, ob ein Maskulinum sich geschlechtsspezifisch auf einen oder mehrere Männer bezieht, oder ob es geschlechtsübergreifend verwendet wird.[11]

Die Annotator*innen wurden instruiert, stets den gesamten Textzusammenhang für die Einordnung zu berücksichtigen. Lautete ein Satz beispielsweise, wie hier aus einem Trainingsdatensatz entnommen: *Die Polizei hatte zuvor zwei mutmaßliche ETA-**Terroristen** in dem Dorf festgenommen. Bei einer Hausdurchsuchung stießen die Ermittler auf das Versteck*, könnte *Terroristen* als GM interpretiert werden, da unklar bleibt, ob es sich ausschließlich um Männer handelt. Wird jedoch im weiteren Textverlauf ergänzt: *Die beiden Festgenommenen sind **Brüder***, sollten die Hilfskräfte diese zusätzliche Information in ihre Annotation einbeziehen und die vorherige Bezeichnung als geschlechtsspezifisches Maskulinum – also als Referenz auf eindeutig männliche Personen – klassifizieren.

Die größte Unsicherheit in der Einordnung von GM besteht bei Personenbezeichnungen im Singular mit unbestimmtem Artikel wie *ein Teilnehmer, ein Chef* oder *ein Manager*. Generell ist anzunehmen, dass im Singular eher Geschlechtsspezifik herrscht (Becker 2008: 66; Kotthoff/Nübling 2024a: 107). Diese Annahme sollten die Hilfskräfte innerhalb eines Textes jeweils kritisch prüfen, u.a. durch Muster der Wiederaufnahme oder auch die Verwendung movierter Formen im gleichen Text, die darauf hindeuten könnte, dass die jeweiligen Autor*innen geschlechtsspezifizierend schreiben. Da diese Interpretationsleistung aber potentiell mit Unsicherheiten behaftet ist, haben wir die Möglichkeit eingeführt, eine Annotation als ‚unsicher' zu markieren. Dies wurde in der Auswertung berücksichtigt.

In einem weiteren Annotationsschritt wurde festgehalten, ob die Personenbezeichnung auf eine konkrete Person oder eine Gruppe von Personen referiert. Entscheidend für die Einordnung war dabei stets der Textkontext: Aus ihm wurde abgeleitet, ob eine identifizierbare Einzelperson oder Gruppe gemeint ist. Ein Satz wie *Ein **Minister** sollte sich immer gut kleiden* wurde beispielsweise als generalisierte Aussage gewertet – vergleichbar mit den Gattungsprädikatoren nach Meineke (2023) – und nicht als Referenz auf eine konkrete Person, da er sich auf die Rolle eines Ministers im Allgemeinen bezieht. Daher wäre Minister hier als GM einzuordnen. Anders verhält es sich bei einer Formulierung wie *Ein **Minister**, der auf dem Podium saß, bestellte einen Cappuccino*. Hier liefert der Relativsatz eine nähere Spezifikation, die darauf hinweist, dass auf eine spezifische männliche Einzelperson referiert wird. Auch in solchen Fällen galt, dass für die Interpretation stets der gesamte Textzusammenhang berücksichtigt werden musste – einschließlich anaphorischer und kataphorischer Verweise. Die folgende Checkliste sollte den Annotator*innen helfen einzuordnen, ob es sich um eine konkrete Person handelt (mehr als ein Kriterium kann zutreffen):

- Eine konkrete Person, auf die sich eine Personenbezeichnung bezieht, ist innerhalb des Textes namentlich genannt.
- Durch kohäsive Mittel der Wiederaufnahme (Proformen, Substitution, Rekurrenz) wird deutlich, dass es sich um eine Einzelperson handelt.
- Die konkrete Person ist aus textexternem Weltwissen ableitbar (*die Bundeskanzlerin, der damalige US-Präsident*).
- Grammatikalische Merkmale weisen auf eine konkrete Person hin (bestimmter Artikel, Possessivpronomen wie *mein*, Wechsel vom unbestimmten zum bestimmten Artikel, Singular, Relativsatz).

---

[11] In unserer Studie konnten keine geschlechtsübergreifenden Feminina identifiziert werden. Alle movierten Formen referieren geschlechtsspezifisch auf Frauen (vgl. Ochs 2025 für eine empirische Analyse generischer Feminina).

Wurde eine Personenbezeichnung als konkrete Referenz eingeordnet, konnte außerdem spezifiziert werden, ob die Geschlechtsidentität der Referenzperson aus dem Kontext erkennbar ist, und wenn ja, welches Geschlecht. Zur Auswahl standen dabei die Einordnungen als *weiblich*, *männlich*, *non-binär* oder *gemischt* (bei Gruppen). Movierungen wurden über die Kategorie ‚moviert' erfasst und korrelierten in unserem Sample immer mit einer geschlechtsspezifisch weiblichen Referenz. In die folgenden Analysen fließen nur übereinstimmende Annotationen ein (vgl. Tabelle 1, letzte Spalte). Der zur Verfügung gestellte Datensatz[12] enthält jedoch alle Items – auch solche, die nur von einer Annotator*in oder nicht übereinstimmend annotiert wurden.

# 4 Ergebnisse

## 4.1 Anteile geschlechtsübergreifender Maskulina

Tabelle 2 zeigt die Anteile der beiden Referenztypen (geschlechtsübergreifend und -spezifisch), unterteilt nach Genus (maskulin/feminin) sowie nach Lexemen. Da keine geschlechtsübergreifenden Feminina vorkommen, entfällt die entsprechende Spalte. Es ist auffällig, dass bei jeder Personenbezeichnung, außer HAMBURGER, der Anteil maskuliner Formen bei über 50% liegt (vgl. Ochs/Rüdiger 2025 für ein ähnliches Ergebnis unter Einbeziehung von Genderzeichen). Insgesamt haben zehn Personenbezeichnungen einen Anteil maskuliner Formen von über 80%. Betrachtet man diese genauer, zeigen sich allerdings erhebliche Unterschiede hinsichtlich des Referenztyps. Während BÜRGER und SPIELER in etwa ähnlich häufig als Maskulina vorliegen (94% und 92% der Gesamtmenge), sind bei BÜRGER 99 % dieser Maskulina geschlechtsübergreifend, bei SPIELER hingegen nur 62%. Noch

---

deutlicher fällt der Unterschied bei Präsident und Mitarbeiter aus: Bei beiden machen Maskulina einen Anteil von 87% aus, von denen bei Mitarbeiter 90% geschlechtsübergreifend sind, bei Präsident hingegen nur 3%. Würde man nur den Anteil oberflächlich maskuliner Formen betrachten, käme man bei diesen Ergebnissen zu dem Schluss, dass Mitarbeiter und Präsident oder Bürger und Spieler sich in etwa gleich verhalten. Nur die manuelle Annotation kann hier zeigen, dass sich die Verwendung der Maskulina als geschlechtsübergreifend oder -spezifisch in den von uns untersuchten Texten sehr stark nach Lexemen unterscheidet. So wäre es 1,6-mal wahrscheinlicher, Bürger als GM anzutreffen als Spieler; bei Mitarbeiter wäre es sogar 31,2-mal wahrscheinlicher als bei Präsident.

| Lexem (stellvertretend für gesamtes Flexionsparadigma) | Gesamt­zahl | Maskulina | | → davon geschlechtsübergreifend | | → davon geschlechtsspezifisch | | Feminina (immer geschlechtsspezifisch) | |
|---|---|---|---|---|---|---|---|---|---|
| | abs | abs | % | abs | % | abs | % | abs | % |
| **Prestigebezeichnungen** | | | | | | | | | |
| BÜRGERMEISTER | 170 | 128 | 75 | 20 | 16 | 108 | 84 | 42 | 25 |
| CHEF | 444 | 356 | 80 | 146 | 41 | 210 | 59 | 88 | 20 |
| MINISTER | 180 | 104 | 58 | 31 | 30 | 73 | 70 | 76 | 42 |
| PFARRER | 115 | 100 | 87 | 13 | 13 | 87 | 87 | 15 | 13 |
| PRÄSIDENT | 358 | 310 | 87 | 9 | 3 | 301 | 97 | 48 | 13 |
| RICHTER | 157 | 137 | 87 | 83 | 61 | 54 | 39 | 20 | 13 |
| **Aktivbezeichnungen** | | | | | | | | | |
| GESCHÄFTSFÜHRER | 174 | 111 | 64 | 10 | 9 | 101 | 91 | 63 | 36 |
| KÜNSTLER | 264 | 202 | 77 | 128 | 63 | 74 | 37 | 62 | 23 |
| LEHRER | 576 | 422 | 73 | 337 | 80 | 85 | 20 | 154 | 27 |
| LEITER | 242 | 171 | 71 | 2 | 1 | 169 | 99 | 71 | 29 |
| MITARBEITER | 611 | 529 | 87 | 478 | 90 | 51 | 10 | 82 | 13 |
| SPIELER | 251 | 231 | 92 | 143 | 62 | 88 | 38 | 20 | 8 |
| **Passivbezeichnungen** | | | | | | | | | |
| BÜRGER | 301 | 283 | 94 | 280 | 99 | 3 | 1 | 18 | 6 |
| FREUND | 1139 | 802 | 70 | 573 | 71 | 229 | 29 | 337 | 30 |
| GASTGEBER | 86 | 61 | 71 | 30 | 49 | 31 | 51 | 25 | 29 |
| SCHÜLER | 573 | 485 | 85 | 436 | 90 | 49 | 10 | 88 | 15 |
| TEILNEHMER | 408 | 365 | 89 | 348 | 95 | 17 | 5 | 43 | 11 |
| **Bevölkerungsbezeichnungen** | | | | | | | | | |
| BERLINER | 56 | 43 | 77 | 24 | 56 | 19 | 44 | 13 | 23 |
| HAMBURGER | 33 | 14 | 42 | 6 | 43 | 8 | 57 | 19 | 58 |
| SCHWEIZER | 46 | 41 | 89 | 19 | 46 | 22 | 54 | 5 | 11 |
| WIENER | 11 | 7 | 64 | 6 | 86 | 1 | 14 | 4 | 36 |

*Tabelle 2: Absolute Mengen und Anteile der Referenztypen nach Lexemen. Maskulina sind grau hinterlegt.*

Hypothese 1 wird damit bestätigt: Die Häufigkeit geschlechtsübergreifender vs. -spezifischer Maskulina variiert deutlich zwischen den Lexemen. Wir halten fest, dass es problematisch ist, von einzelnen Lexemen Rückschlüsse auf Verteilungen in der gesamten Population der Personenbezeichnungen zu ziehen. Fallanalysen anhand einzelner, nicht näher klassifizierter Lexeme (z.B. bei Adler/Hansen 2020; Krome 2021) müssen auch als solche verstanden werden und sollten nicht zu generalisierten Aussagen über Frequenzverteilungen führen. Wir diskutieren in Abschnitt 5, welche möglichen Fragen sich an diese sehr ungleichen Verteilungen anschließen, z.B. hinsichtlich der kognitiven Speicherung und der Verwendung genderinklusiver Sprache.

Zur Überprüfung der zweiten Hypothese (H2) zeigen wir in Abbildung 2 zunächst die deskriptiven Ergebnisse der Annotationen. Wir sehen bei der lexemspezifischen Verteilung nach Bezeichnungskategorie deutliche Unterschiede in den Kategorien geschlechtsübergreifendes Maskulinum, geschlechtsspezifisches Maskulinum und geschlechtsspezifisches Femininum. Es deutet sich an, dass Prestigebezeichnungen häufiger im spezifischen Maskulinum vorkommen als Passivbezeichnungen, bei denen Vorkommen im GM überwiegen. Aktiv- und Bevölkerungsbezeichnungen zeigen kein klares Muster. Bei Bevölkerungsbezeichnungen ist außerdem zu beachten, dass sehr viel weniger Fälle in die Analysen eingehen.

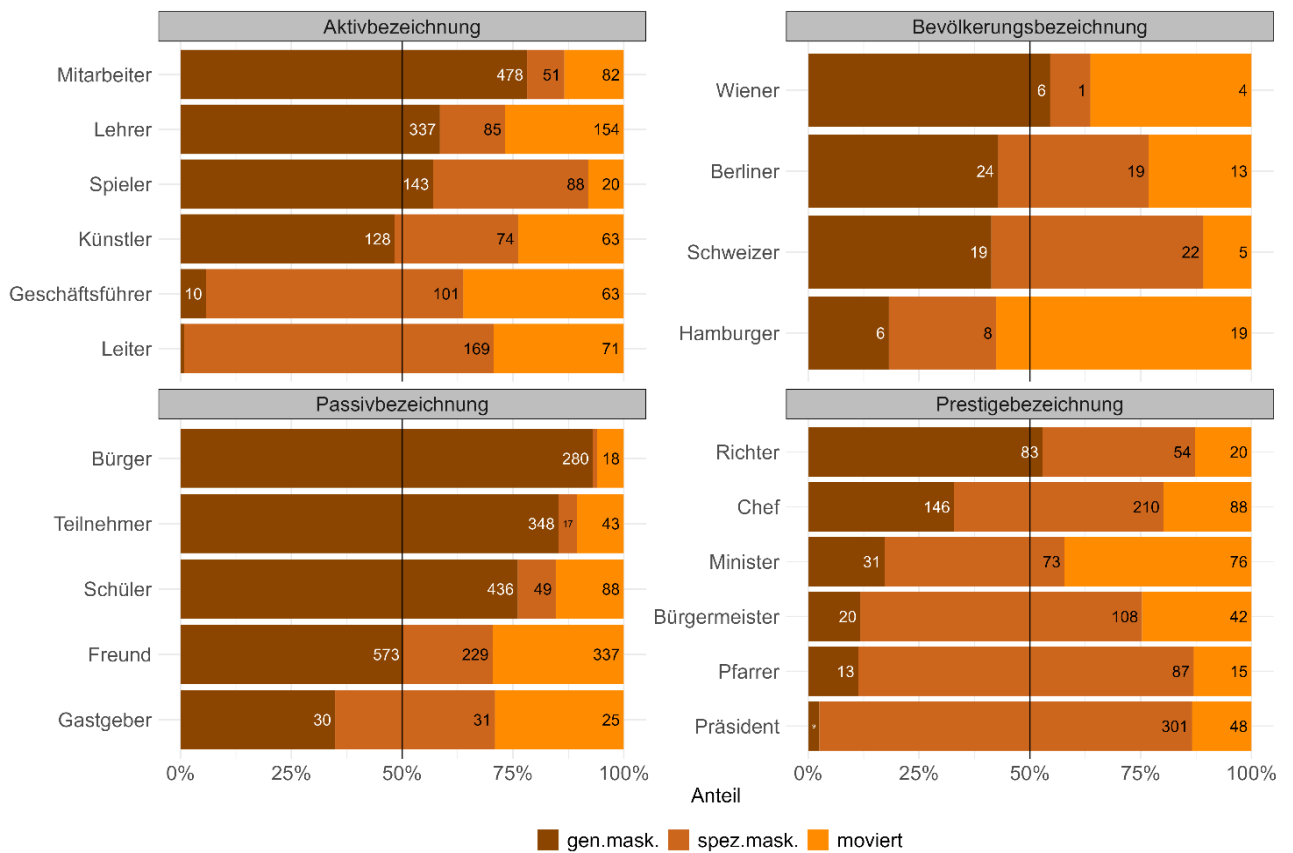

*Abbildung 2: Lexemspezifische Verteilung der Kategorien geschlechtsübergreifendes Maskulinum, geschlechtsspezifisches Maskulinum und geschlechtsspezifisches Femininum (= moviert), geordnet nach den Bezeichnungsarten Aktiv-, Passiv-, Bevölkerungs- und Prestigebezeichnung. Die Zahlen in den Balken stellen absolute Häufigkeiten dar. Die maskulinen Formen auf den y-Achsen stehen für das gesamte Flexionsparadigma.*

Deutliche Unterschiede zeigen sich bei den Bezeichnungskategorien auch dahingehend, welche Geschlechtsidentität bei einer konkreten Referenz aus dem Textkontext erkennbar ist. Bei Prestigebezeichnungen wird in 58% aller Fälle auf eine konkrete männliche Person referiert, bei

Aktivbezeichnungen in 26 %, dagegen bei Passivbezeichnungen nur in 13% der Fälle (vgl. Abb. 3).[13] Es zeigt sich also eine Dominanz männlicher Referenzpersonen bei prestigeträchtigen Berufen und Rollen (vgl. ein ähnliches Ergebnis bei Müller-Spitzer et al. 2024), während bei Aktivbezeichnungen der Geschlechteranteil ausgewogener, aber dennoch leicht männlich dominiert ist. Ob dies allein den gesellschaftlichen Hintergrund widerspiegelt, dass noch immer mehr Männer prestigeträchtige Rollen innehaben, oder ob zusätzlich ein Selektions-Bias in Pressetexten wirkt, bei dem in der Darstellung dieser Rollen verstärkt auf konkrete Männer zurückgegriffen wird, muss künftige Forschung klären.[14]
Es zeigt sich auch deutlich, dass Passivbezeichnungen im Großteil der Fälle (68%) nicht auf konkrete Personen referieren; bei Aktivbezeichnungen ist der Anteil nicht-konkreter Referenzen (54%) ebenfalls größer als der der konkreten.

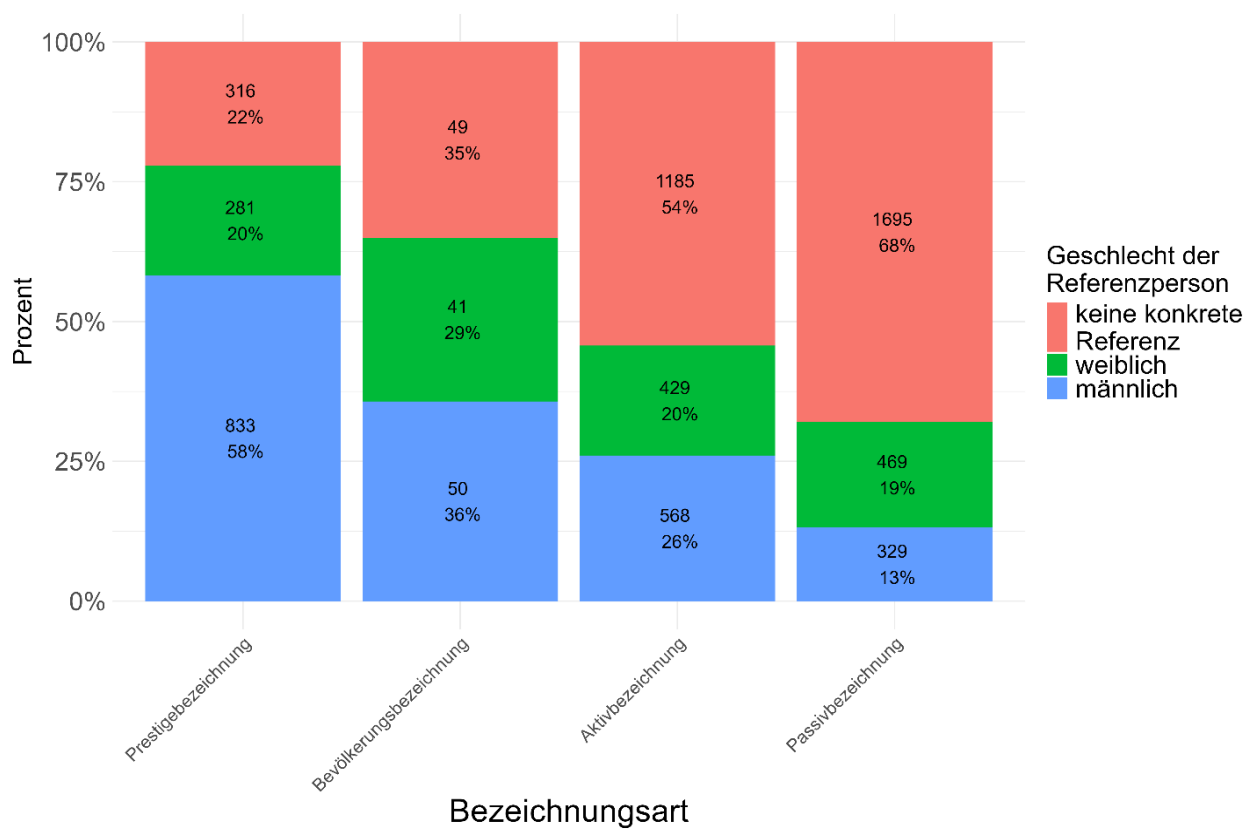

*Abb. 3: Anteile der Referenzen auf konkret männliche und weibliche Personen nach Bezeichnungskategorie (non-binär kam im Sample nicht vor).*

Um zu überprüfen, ob sich die Bezeichnungskategorien hinsichtlich der Verteilung von geschlechtsübergreifenden und geschlechtsspezifischen Maskulina statistisch signifikant unterscheiden, wird ein Pearson-Chi-Quadrat-Test durchgeführt. Konkret wird kontrolliert, ob

---

[13] Obwohl *non-binär* als Annotationskategorie existierte, gab es keine entsprechenden doppelten Annotationen.
[14] In einem neuen Forschungsprojekt „GENELLI: Am Empirical Look into the Language-Cognition-Interface" beschäftigen wir uns u.a. genau mit dieser Frage (https://www.ids-mannheim.de/lexik/pb-l3/empirische-genderlinguistik/genelli/).

Personenbezeichnungen aus den Kategorien Aktiv- und Prestigebezeichnung überwiegend als spezifische Maskulina (H2.1) und die Bezeichnungen aus den Kategorien Bevölkerungs- und Passivbezeichnung vorwiegend als geschlechtsübergreifendes Maskulinum verwendet werden (H2.2). Der Pearson-Chi-Quadrat-Test dient dazu, zu überprüfen, ob ein signifikanter Zusammenhang zwischen den Bezeichnungskategorien (Aktiv-, Passiv-, Bevölkerungs- und Prestigebezeichnungen) und der Form (geschlechtsübergreifendes Maskulinum, geschlechtsspezifisches Maskulinum bzw. Femininum) besteht. Es ergibt sich ein signifikanter Zusammenhang zwischen den beiden Variablen, $\chi^2$(6, N = 6195) = 1035,5, $p$ < 0,001. Zur Beurteilung der Effektstärke wurde der Cramér's V-Wert berechnet, der mit 0,29 auf einen mittelstarken Zusammenhang hindeutet.[15] Bezeichnungsart und Form der Referenz hängen somit zwar statistisch signifikant zusammen, der Zusammenhang weist jedoch nur eine moderate Effektstärke auf. Der Mosaikplot in Abbildung 4 veranschaulicht zusätzlich die Richtung der Zusammenhänge: Die blauen Flächen zeigen an, dass eine Kombination signifikant häufiger als erwartet vorkommt, während die roten Flächen auf Kombinationen hinweisen, die signifikant seltener auftreten. Die erwarteten Häufigkeiten ergeben sich aus den Grundwahrscheinlichkeiten der Form und der Bezeichnungsart. Die Fläche der Rechtecke bildet die absoluten Häufigkeiten der Kombinationen ab. Passiv- und Prestigebezeichnungen zeigen jeweils entgegengesetzte Muster: Bei Passivbezeichnungen treten GM signifikant häufiger auf, während bei Prestigebezeichnungen geschlechtsspezifische Maskulina häufiger vorkommen als erwartet. Aktivbezeichnungen weisen keine signifikanten Abweichungen von der Erwartung auf. Bei den Bevölkerungsbezeichnungen zeigt sich, entgegen H2.2, dass geschlechtsübergreifende Maskulina etwas seltener sind als erwartet – für spezifische Maskulina ist hier kein Zusammenhang zu verzeichnen.

---

[15] Cohen (1988) folgend werden Werte von *V* > 0,1 als schwacher Effekt, *V* > 0,3 als moderater und *V* > 0,5 als starker Effekt interpretiert.

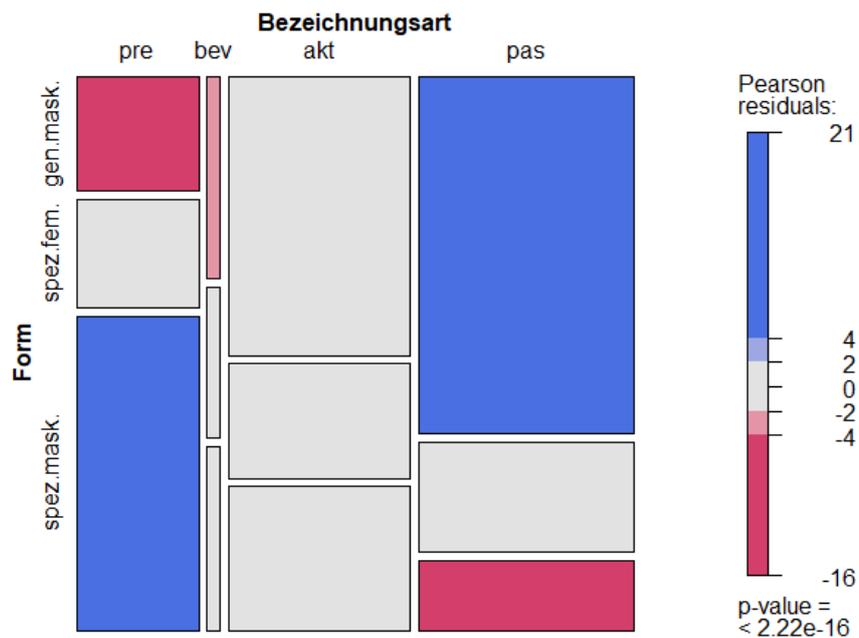

*Abbildung 4: Mosaikplot zur Visualisierung des Chi-Quadrat-Tests für die Variablen ‚Bezeichnungsart' (pre = Prestige, bev = Bevölkerung, akt = Aktiv, pas = Passiv) und ‚Form' (spez.mask. = geschlechtsspezifisches Maskulinum, spez.fem. = geschlechtsspezifisches Femininum, gen.mask. = GM)*

Hypothese 2, die annimmt, dass sich die verschiedenen Bezeichnungsarten hinsichtlich der Verwendung von GM unterscheiden, kann also nur bedingt akzeptiert werden. Unterhypothese H2.1 kann nur teilweise bestätigt werden, da ein signifikanter Zusammenhang zugunsten spezifischer Maskulina lediglich bei Prestigebezeichnungen nachweisbar ist. Für Aktivbezeichnungen zeigt sich kein solcher Zusammenhang. Ähnlich verhält es sich mit Unterhypothese H2.2: Die gemäß dieser Hypothese erwarteten Zusammenhänge treten nur bei Passivbezeichnungen, jedoch nicht bei Bevölkerungsbezeichnungen auf. Letzteres könnte jedoch durch die geringen Belegmengen bei den Bevölkerungsbezeichnungen beeinflusst sein. Da nach dem Aussortieren unpassender Texte kein Nachsampling durchgeführt wurde, kann nicht sichergestellt werden, dass für die Bevölkerungsbezeichnungen alle drei Referenztypen in ausreichender Weise vorliegen. Dennoch lassen sich aus den Daten zwei Pole ableiten (Prestige vs. Passiv), mit entsprechenden Grauzonen im Bereich der Aktivbezeichnungen. Allerdings müssen diese Resultate vor dem Hintergrund interpretiert werden, dass die Einteilung in die einzelnen Bezeichnungsarten – wie anfangs herausgestellt wurde – nicht immer eindeutig ist. Insgesamt zeigt sich jedoch, dass die Lexemspezifik eine zentrale Rolle spielt. Unsere Ergebnisse unterstreichen somit, dass sprachliche Muster grundsätzlich stark kontextsensitiv modelliert werden müssen, auch weil sich innerhalb der Bezeichnungskategorien einige Variation zeigt.

## 4.2   Einfluss von Numerus und Definitheit

Abbildungen 5 und 6 visualisieren zunächst die Verteilung nach Referenztyp, Genus und Numerus (Abb. 5) bzw. nach Definitheit (Abb. 6). In diesen Übersichten sollten wieder die Bevölkerungsbezeichnungen mit Vorsicht interpretiert werden, da sie in einigen Kategorien sehr wenige Belege aufweisen.[16] Es ist direkt ersichtlich, dass sich die einzelnen Lexeme insgesamt sehr deutlich hinsichtlich dieser Verteilungen unterscheiden.

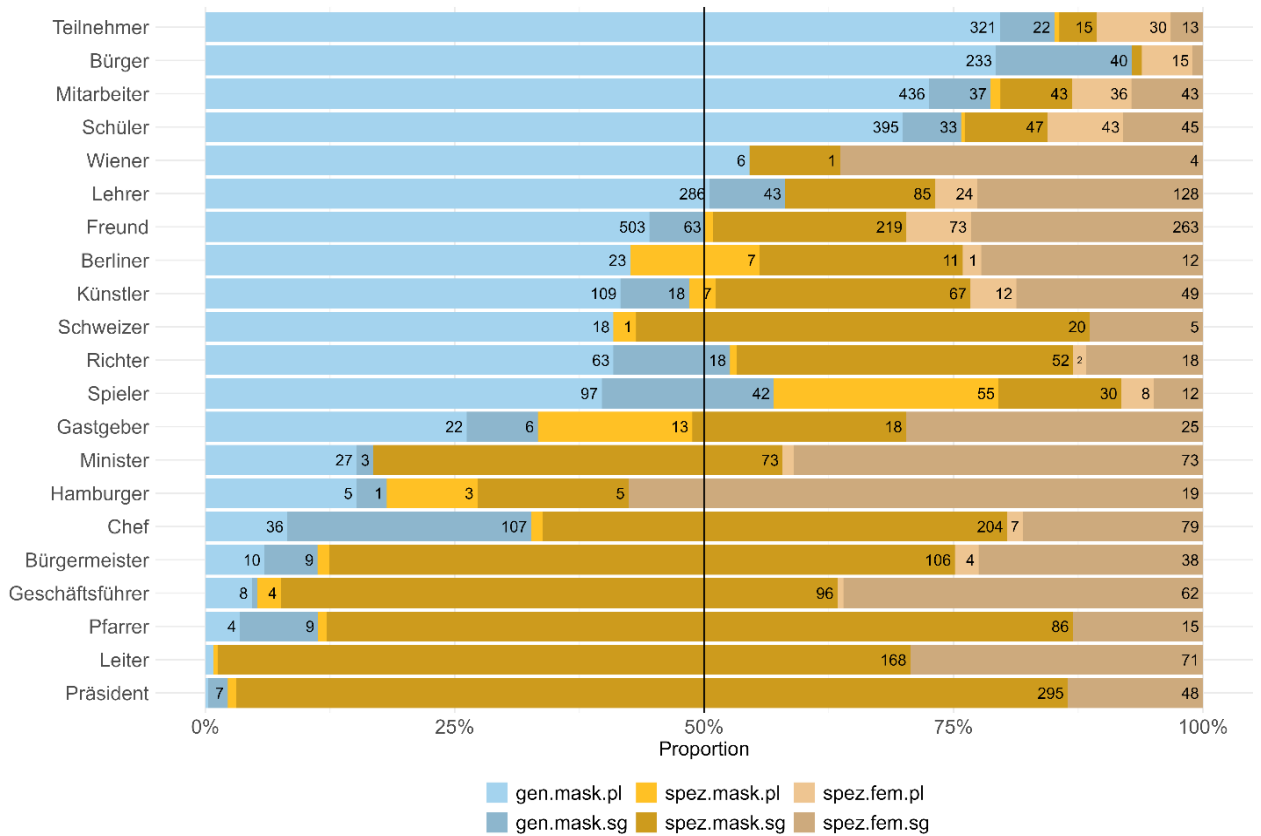

*Abbildung 5: Verteilung geschlechtsübergreifender Maskulina und geschlechtsspezifischer Maskulina und Feminina nach Lexemen, zusätzlich aufgeschlüsselt nach Numerus (absteigend sortiert nach Anteil geschlechtsübergreifender Maskulina im Plural). Die maskulinen Formen auf der y-Achse stehen für das gesamte Flexionsparadigma.*

---

[16] Zum Beispiel gibt es für Wiener in der Kategorie *geschlechtsspezifisches Maskulinum Singular* nur einen Beleg, ebenso für Hamburger in der Gruppe *geschlechtsübergreifendes Maskulinum Plural* (siehe Abb. 5).

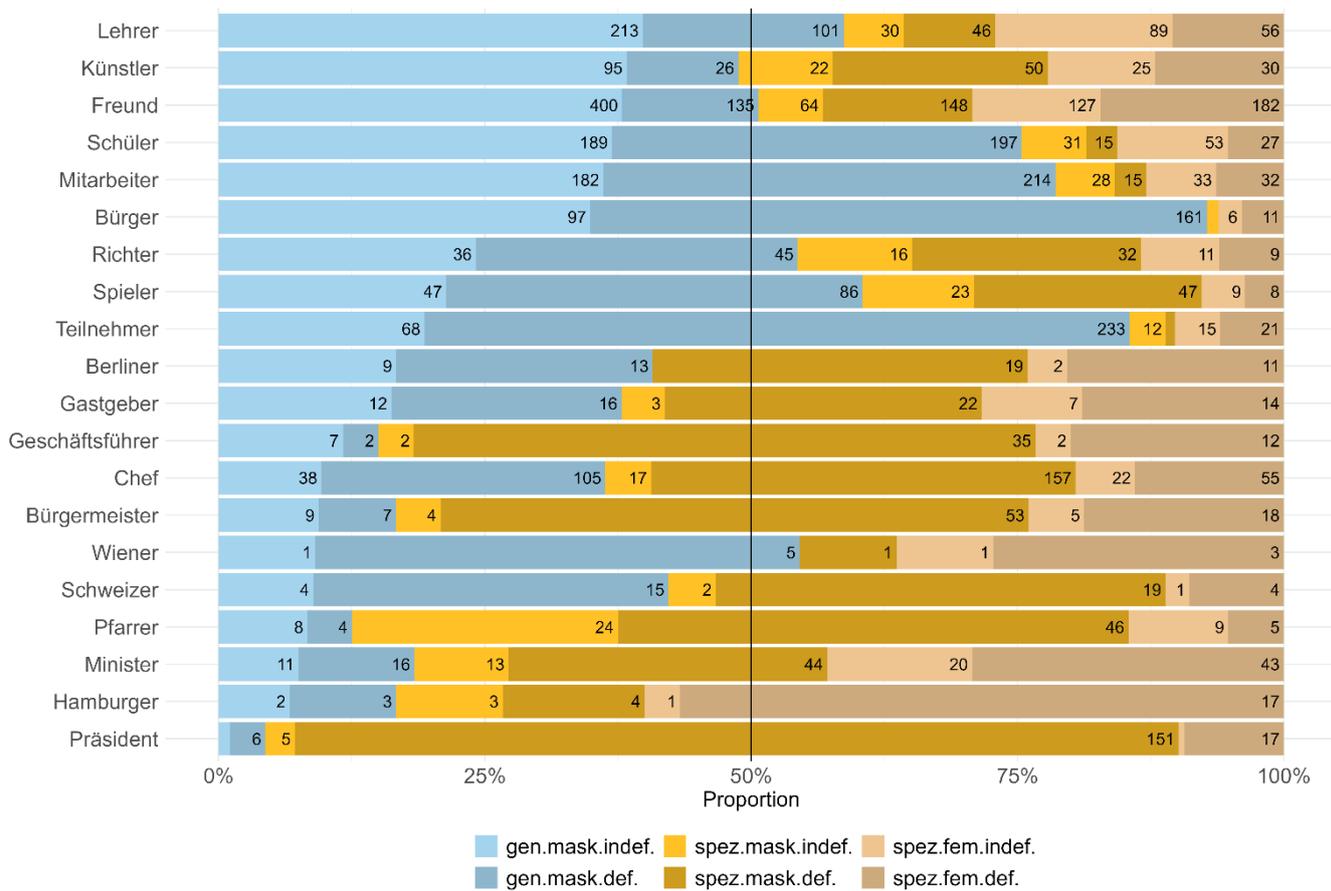

*Abbildung 6: Verteilung geschlechtsübergreifender Maskulina und geschlechtsspezifischer Maskulina und Feminina nach Lexemen, zusätzlich aufgeschlüsselt nach Definitheit (absteigend sortiert nach Anteil geschlechtsübergreifender Maskulina in indefiniten NPs). Die maskulinen Formen auf der y-Achse stehen für das gesamte Flexionsparadigma.*

Die Hypothesen 3 und 4 wurden ebenfalls mit Pearson-Chi-Quadrat-Test überprüft. Zunächst wurde getestet, ob ein signifikanter Zusammenhang zwischen den Kategorien von *Numerus* (Singular, Plural) und *Form* besteht. Der Test ergab einen signifikanten Zusammenhang, $\chi^2$(2, N = 6114) = 3270,2, p < 0,001. Der Cramér's V-Wert von 0,73 zeigt einen starken Effekt. Abbildung 7 verdeutlicht die Richtung des Zusammenhangs: GM kommen signifikant häufiger als erwartet im Plural vor (H3.1), die beiden spezifischen Formen wiederum signifikant häufiger im Singular (H3.2). Die beiden Unterhypothesen bestätigen sich somit.

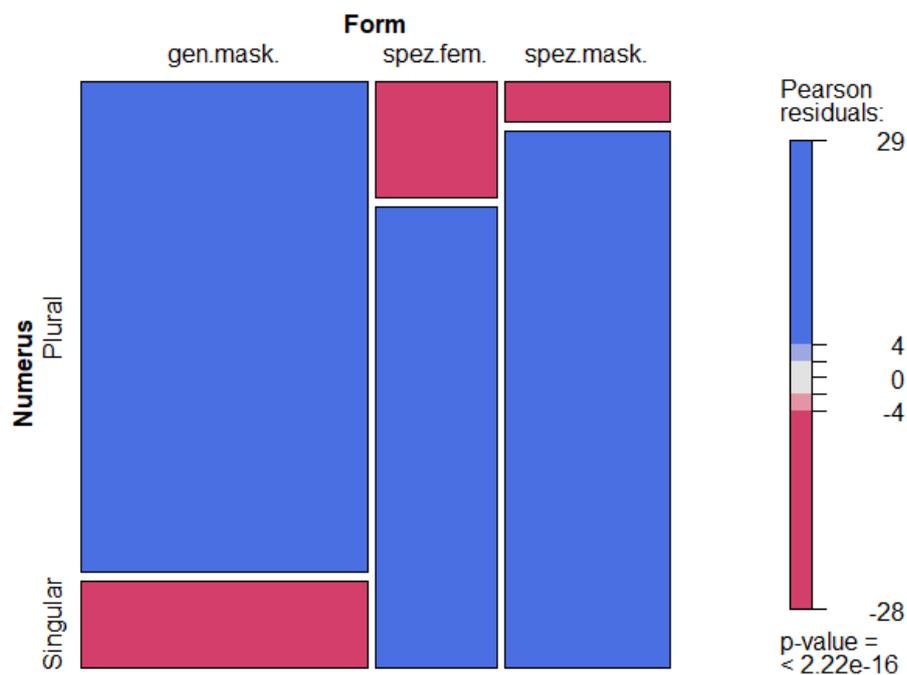

*Abbildung 7: Mosaikplot zur Visualisierung des Chi-Quadrat-Tests für die Variablen ‚Form' und ‚Numerus'.*

Der Test für die Kategorien *Definitheit* (definit, indefinit) und *Form* ist ebenfalls statistisch signifikant, $\chi^2$(2, N = 5125) = 253,79, *p* < 0,001. Der Cramér's V-Wert von 0,22 zeigt allerdings, dass es sich nur um einen schwachen Effekt handelt, d.h. die untersuchten Variablen sind nur in geringem Maße miteinander verbunden, auch wenn der Zusammenhang signifikant ist.

Abbildung 8 verdeutlicht die Richtung des Zusammenhangs für die Definitheit: GM kommen signifikant häufiger als erwartet in indefiniter Form vor. Spezifische Maskulina treten signifikant häufiger als erwartet in definiter Form auf. Die Hypothese 4 (signifikanter Zusammenhang von Definitheit und Referenzialitätstyp) und die Unterhypothese 4.1 (GM überwiegend in indefiniter Form) bestätigen sich somit. Dieser Zusammenhang zeigt sich bei spezifischen Feminina nicht, d.h. die Unterhypothese 4.2, dass alle spezifischen Formen signifikant mit Definitheit assoziiert sind, bestätigt sich nur bedingt.

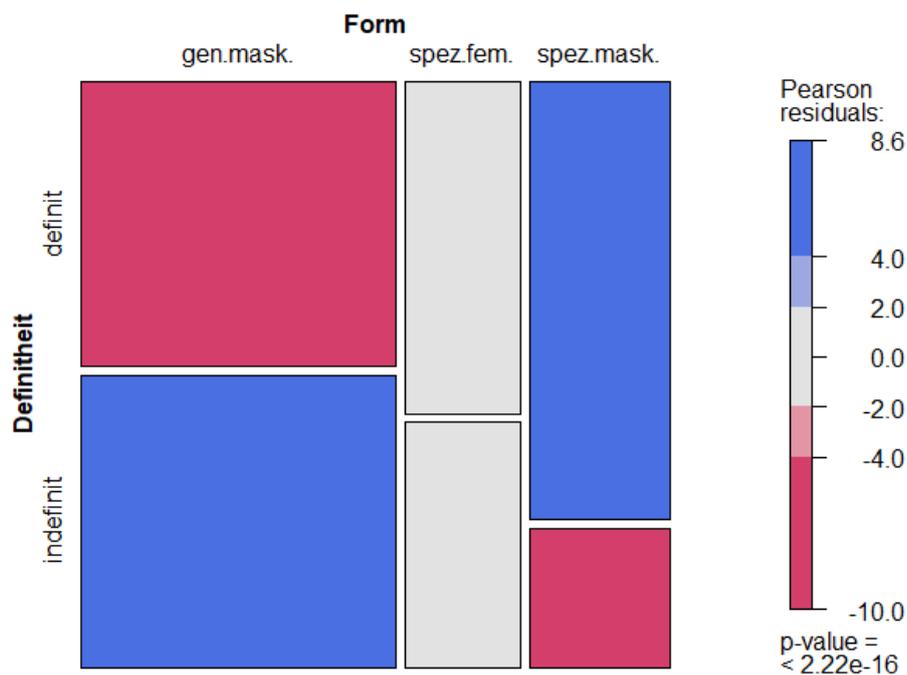

*Abbildung 8: Mosaikplot zur Visualisierung des Chi-Quadrat-Tests für die Variablen ‚Form' und ‚Definitheit'.*

Auch wenn die statistischen Auswertungen die Hypothesen für Gruppen von Personenbezeichnungen in Teilen bestätigen, zeigen insbesondere die Abbildungen 5 und 6, dass es erhebliche lexemspezifische Abweichungen von diesen Trends gibt. So weisen beispielsweise CHEF oder SPIELER einen erhöhten Anteil an GM im Singular auf, während das GM bei BÜRGER oder TEILNEHMER besonders häufig in definiten Nominalphrasen vorkommt. Eine lexemspezifische Analyse ist daher auch hier sinnvoll, und Aussagen über die ‚Typizität' von Verwendungsformen können über Personenbezeichnungen als Gesamtklasse kaum getroffen werden.

## 4.3    Geschlechtsübergreifende Maskulina als Gattungsbezeichnungen

Wir haben in den vorherigen Abschnitten Annahmen über das Vorkommen von GM hinsichtlich Numerus und Definitheit überprüft. Nun widmen wir uns noch der Frage, ob sich Meinekes (2023: 55) Annahme, dass eine „Hauptdomäne" der geschlechtsübergreifenden Maskulina Gattungsbezeichnungen sind, anhand unserer Daten bestätigen lässt. Wir interpretieren ‚Hauptdomäne' dabei im Sinne von ‚besonders häufig und dominant' und prüfen, ob sich diese Annahme in unseren Daten widerspiegelt. Um dies zu überprüfen, wurden die 2.862 Tokens aus dem Dataset, die von beiden Annotator*innen als GM annotiert wurden, anhand des Kontextes weiter differenziert. Dabei wurde geprüft, ob ein Beleg als Gattungsprädikator (nach Meinekes Einordnung) verwendet wird. Zusätzlich wurde festgehalten, ob die Einordnung eindeutig oder unsicher war. Diese Annotation wurde nicht doppelt, sondern nur von einer Person vorgenommen. Die Einstufung als Gattungsbezeichnung erfolgte, wenn eine Gruppe als Ganzes gemeint ist bzw. alle Mitglieder einer Gattung (vgl. dazu Kotthoff/Nübling 2024a: 103–104). Die folgenden Beispiele aus unseren Daten illustrieren diese Verwendungsweise des GM:

- *Die **Schweizer** sind gar nicht so, wie man immer sagt.*
- ***Künstler** sehen die Welt durch eine andere Linse, sie können Themen direkter aufgreifen als viele andere.*
- *Gute **Gastgeber** haben immer das Wohl der Gäste im Blick.*

Im letzten Beleg wird zwar nicht auf die gesamte Klasse referiert, sondern nur auf den Teil der *guten Gastgeber*. Solche Belege wurden trotzdem als Gattungsbezeichnungen eingeordnet, allerdings als *unsicher* klassifiziert.

Im Ergebnis wurden von den 2.862 nur 184 Belege (6%) als Gattungsprädikatoren klassifiziert, von denen wiederum 73 (40%) als unsicher eingestuft wurden.[17] Diese Ergebnisse legen nahe, dass die Verwendung als Gattungsprädikatoren nicht als Hauptdomäne des GM betrachtet werden kann. Hypothese H5 bestätigt sich damit nicht. Gleichzeitig verdeutlichen diese Nachannotationen, dass selbst einfache Klassifikationssysteme wie die binäre Kategorisierung nach Gattungsprädikator (ja/nein) bei der Einordnung von authentischen Belegen aus dem Sprachgebrauch erhebliche Unsicherheiten erzeugen.

Insgesamt zeigt sich in der vorliegenden Studie – wie auch in anderen Debatten um das GM (vgl. u.a. Müller-Spitzer et al. 2024) –, dass es problematisch ist, Aussagen über diesen Gegenstandsbereich ohne empirische Untersuchungen des Sprachgebrauchs zu treffen. Zudem wird die begriffliche Unschärfe des Objekts deutlich: Die etablierte Bezeichnung *generisches Maskulinum* vermischt das referenzsemantische Konzept der generischen Referenz und die pragmatische, geschlechtsübergreifende Verwendung von Maskulina. Diewald (2025a: 20–22) weist daher darauf hin, dass allein die Benennung *generisches Maskulinum* zu vielen Missverständnissen führt, da generische, klassenbezogene Referenz (wie bei Meinekes Gattungsprädiaktoren) nur eine von mehreren Möglichkeiten ist, Maskulina geschlechtsübergreifend zu verwenden.

---

[17] Bei den Belegen, die nicht als Gattungsprädikatoren eingeordnet wurden (*N* = 2.678) liegt der Anteil unsicherer Klassifizierungen bei 6,8%.

## 5   Diskussion und Ausblick

Unsere empirischen Untersuchungen zeigen, dass eine umfangreiche manuelle Annotation großer Textmengen verlässliche Daten zum tatsächlichen Vorkommen geschlechtsübergreifender Maskulina in Pressetexten liefert. Dadurch lassen sich bisherige Vermutungen teils präzisieren, teils widerlegen. In der Literatur verbreitete verabsolutierte Aussagen wie *kommt ausschließlich/überwiegend als XY vor* lassen sich durch unsere Daten nicht bestätigen. Stattdessen zeigt sich eine große Vielfalt an möglichen Formen und Ausprägungen. Dennoch treten statistisch signifikante Zusammenhänge zwischen dem GM und den Faktoren Bezeichnungsart, Numerus und Definitheit auf. Die zentralen Ergebnisse unserer Untersuchung lassen sich wie folgt zusammenfassen:

(1)  Häufigkeitsaussagen zum GM im Sprachgebrauch sind nur möglich, wenn verschiedene Lexeme berücksichtigt werden, da die Verteilungen auf lexikalischer Ebene stark variieren.

(2)  Klassen von Personenbezeichnungen unterscheiden sich nachweisbarei der Verwendung geschlechtsübergreifender und -spezifischer Formen: Passive Rollen wie Teilnehmer oder Schüler werden häufiger geschlechtsübergreifend verwendet, während Prestigebezeichnungen wie Chef oder Manager meist spezifisch männlich referieren.

(3)  Das GM tritt überwiegend im Plural auf.

(4)  Das GM kommt hauptsächlich in indefiniten Nominalphrasen vor.

(5)  Das GM wird nicht vorrangig zur Bezeichnung einer Klasse oder Gattung verwendet.

Wir möchten in diesem Diskussionsteil verschiedene Anschlussfragen aufzeigen, die sich daraus ergeben und in weiteren Studien untersucht bzw. berücksichtigt werden könnten. Eine eher qualitativ ausgerichtete korpuslinguistische Vertiefung der vorliegenden Studie wäre denkbar, indem der Kontext der Verwendungsweisen der Lexeme genauer analysiert wird. So ließe sich ein noch stärker auf Kontextsensitivität fokussiertes Verständnis des Phänomens GM gewinnen.

Für psycholinguistische Studien können die vorliegenden Ergebnisse dazu dienen, sprachliches Stimulusmaterial besser am Sprachgebrauch auszurichten. In Satzfortsetzungsstudien nach dem Paradigma von Gygax et al. (2008) wurden lange ausschließlich GM in definiten Nominalphrasen verwendet. Körner et al. (2022) dagegen integrierten in ihre Studie auch indefinite Nominalphrasen mit *alle*, z.B. *Alle Schwimmer warteten in einem Raum*[18]. Auf Basis unserer Daten könnte nun noch besser beachtet werden, welche Lexeme in welcher Art der Einbettung vorkommen, sodass die Stimulussätze diese Muster abbilden können.   Außerdem erscheint es uns sinnvoll, die lexemspezifische Vorkommenshäufigkeit geschlechtsübergreifender vs. -spezifischer Maskulina und Feminina als Faktor in Analysen mit einzubeziehen. So könnte geprüft werden, ob Lexeme, die im Sprachgebrauch vorwiegend als GM vorkommen, in psycholinguistischen Studien einen schwächeren *male bias* zeigen als solche, die vorwiegend als geschlechtsspezifische Maskulina verwendet werden oder deren movierte Form besonders häufig ist (vgl. dazu Backer/Cuypere 2012).

---

[18] S. https://osf.io/zm27r.

Außerdem wäre es interessant zu kontrollieren, ob die von uns beobachteten Verteilungen mental repräsentiert sind. Konkret könnte man prüfen, ob die lexemspezifische Wahrscheinlichkeit, mit der ein Maskulinum als GM verwendet wird, erlernt und kognitiv zugänglich ist. Psycholinguistische Studien könnten testen, ob Proband*innen die ‚wahrscheinliche' Form eines Maskulinums – geschlechtsübergreifend oder -spezifisch – je nach lexikalischem Kontext abrufen können und ob sich die Erwartbarkeit experimentell beobachten lässt. Wenn dies zutrifft, könnte dies Implikationen für die Verwendung genderinklusiver Sprache haben. Eine mögliche These wäre, dass bei Personenbezeichnungen, die häufig als geschlechtsspezifisches Maskulinum verwendet werden, eine sichtbare genderinklusive Umformulierung besonders wichtig ist, um zu verhindern, dass die erlernte spezifische Lesart die intendierte geschlechtsübergreifende Lesart ‚überschreibt'.

In diesem Sinne könnte man argumentieren, dass zur Umleitung der Standardinterpretation – z.B. dass das Wort PRÄSIDENT in den meisten Fällen als geschlechtsspezifisches Maskulinum verwendet wird – eine Form wie *Präsident*in* oder eine Doppelform wie *Präsidentin oder Präsident* notwendig ist, um explizit auf die unwahrscheinlichere, geschlechtsübergreifende Intention hinzuweisen (sofern diese intendiert ist). Bei einem Lexem wie BÜRGER oder EINWOHNER wäre es dagegen weniger wichtig, explizit genderinklusive Formen zu verwenden, da diese Maskulina häufig geschlechtsübergreifend verwendet werden, was die geschlechtsübergreifende Lesart von vornherein nahelegt. Hintergrund dieser Überlegung ist die Annahme, dass Sprecher*innen für vorhersehbare Bedeutungen kurze Formen oder Nullkodierungen verwenden können, während für unvorhersehbare Bedeutungen ein größerer Kodierungsaufwand erforderlich ist (Haspelmath 2021: 624). Diese Erklärung geht von hoher Frequenz zu Vorhersagbarkeit und von Vorhersagbarkeit zu ‚short coding' (vgl. Haspelmath 2021: 607). Daraus lässt sich umgekehrt ableiten, dass die Notwendigkeit einer expliziten Kodierung für eine geschlechtsübergreifende Intention steigt, wenn ein Maskulinum bei bestimmten Personenbezeichnungen überwiegend geschlechtsspezifisch verwendet wird. Hier wäre eine zusätzliche Kodierung als Hinweis besonders wichtig, um gegen die lexemspezifisch ‚übliche' Lesart des Maskulinums zu wirken.[19] Andererseits zeigen Daten zu genderinklusiven Formulierungen – etwa

---

[19] Als weiterer Diskussionspunkt sind auch mögliche Implikationen für das Konzept der *Markiertheit* zu erwähnen. In der Linguistik wird bereits seit Langem über Markiertheit diskutiert (Diewald 2025a; für einen Überblick s. Haspelmath 2006). Ein Problem bei der Modellierung von Markiertheit und Nichtmarkiertheit sprachlicher

Doppelformen (Müller-Spitzer et al. 2025) oder Genderzeichen (Ochs/Rüdiger 2025) –, dass diese insbesondere solche Lexeme betreffen, die in unserer Studie vermehrt geschlechtsübergreifend verwendet werden oder als Passivbezeichnungen gelten würden (z. B. BÜRGER, MIETER, ANWOHNER). Dies könnte darauf hindeuten, dass Personenbezeichnungen, die besonders häufig im GM vorkommen, ein hohes ‚geschlechtsübergreifendes Potenzial' haben, das durch neue genderinklusive Formen aufgegriffen wird. Wir gehen davon aus, dass beide Interpretationslinien für den genderinklusiven Sprachgebrauch relevant sind – je nachdem, welches Ziel mit den Ersatzformen in einem bestimmten Kontext verfolgt wird. So kann es etwa darum gehen, eine direkte Anrede zu ermöglichen (*Liebe Bürgerinnen und Bürger*) oder hervorzuheben, dass eine große Bevölkerungsgruppe unabhängig von Geschlechtsidentitäten gemeint ist (*alle Anwohner\*innen*). Gleichzeitig kann gerade in appellativen Textsorten wie Stellenanzeigen das Aufbrechen geschlechtsspezifischer Lesarten eine tragende Rolle spielen (vgl. z.B. Hetjens/Hartmann 2024), wie wir es exemplarisch für PRÄSIDENT erläutert haben. Je mehr also über die lexemspezifischen Verteilungen bekannt ist, desto besser kann abgewogen werden, welche Arten genderinklusiver Ersatzformen je nach Kommunikationsziel notwendig und angemessen sein können.

Auf Grundlage unserer Studie, die unserer Kenntnis nach erstmals das GM in diesem Umfang empirisch im Sprachgebrauch untersucht, können somit interessante Folgestudien konzipiert werden. Diese könnten uns nicht nur feinkörnigeres Wissen über die lexemspezifische Verteilung verschiedener Arten von Maskulina im Sprachgebrauch liefern, sondern auch die Implikationen dieser Verteilungen aufzeigen.

# 6   Datenverfügbarkeit

Das Datenset, auf dem die Studie basiert, ist zusammen mit einer Readme-Datei im OSF verfügbar.

---

Ausdrücke ist, dass zwar viele einprägsame Beispiele existieren, aber oft unklar bleibt, was genau Markiertheit bedeutet und wie sie empirisch erfasst werden kann. Haspelmath (2021: 623) argumentiert daher, dass Markiertheit als universelles Konzept eine problematische Analysekategorie darstellt und dass viele der weit verbreiteten und systematischen Kodierungsmuster in den Sprachen der Welt auf gut verstandene funktionale und adaptive Kräfte zurückzuführen sind (Haspelmath 2021: 605). Es bestünde demnach keine Notwendigkeit, auf das Konzept der Markiertheit oder andere schwer fassbare Mechanismen zurückzugreifen. Wenn man also ‚Un-Markiertheit' durch ‚Üblichkeit' oder ‚Häufigkeit' ersetzen würde, unterstützen unsere Daten die Annahme, dass Markiertheit nicht (nur) grammatisch kodiert ist (z. B. dass das Maskulinum die unmarkierte Form sei, weil es die Derivationsbasis darstellt), sondern gerade bei komplexen Referenzphänomenen auch lexemspezifisch und pragmatisch betrachtet werden muss. So wäre für Personenbezeichnungen wie PRÄSIDENT oder GESCHÄFTSFÜHRER das geschlechtsspezifische Maskulinum die ‚unmarkierte' (da häufigere) Form, während für BÜRGER oder TEILNEHMER das geschlechtsübergreifende Maskulinum als unmarkiert (da häufiger) betrachtet werden könnte.

# 7 Literatur

Acke, Hanna (2019): Sprachwandel durch feministische Sprachkritik. In: Zeitschrift für Literaturwissenschaft und Linguistik 49(2), S. 303–320. https://doi.org/10.1007/s41244-019-00135-1.

Adler, Astried/Hansen, Karolina (2020): *Studenten, Studentinnen, Studierende*? Aktuelle Verwendungspräferenzen bei Personenbezeichnungen. In: Muttersprache 130(1), S. 47–63.

Backer, Maarten De/Cuypere, Ludovic De (2012): The Interpretation of Masculine Personal Nouns in German and Dutch: A Comparative Experimental Study. In: Language Sciences 34(3), S. 253–268. https://doi.org/10.1016/j.langsci.2011.10.001.

Becker, Thomas (2008): Zum generischen Maskulinum: Bedeutung und Gebrauch der nicht-movierten Personenbezeichnungen im Deutschen. In: Linguistische Berichte 213, S. 65–76.

Bröder, Hannah-Charlotte/Rosar, Anne (2025): Welche Geschlechtsvorstellung erzeugen nicht-referenzielle (generische) Maskulina? Zum Wechselspiel von Stereotypen und Grammatik. In: Linguistik Online 140(8). https://doi.org/10.13092/gatha032.

Bühlmann, Regula (2002): Ehefrau Vreni haucht ihm ins Ohr... Untersuchung zur geschlechtergerechten Sprache und zur Darstellung von Frauen in Deutschschweizer Tageszeitungen. In: Linguistik Online 11(2). https://doi.org/10.13092/lo.11.918.

Bülow, Lars/Jakob, Katharina (2017): Genderassoziationen von Muttersprachlern und DaF-Lernern – grammatik- und/oder kontextbedingt? In: Osnabrücker Beiträge zur Sprachtheorie S. 137–163.

Cohen, Jacob (1988): Statistical Power Analysis for the Behavioral Sciences. 2. Auflage. New York, NY: Psychology Press.

Corbett, Greville G. (2013): Introduction. In: Corbett, Greville G. (Hrg.): The Expression of Gender. De Gruyter Mouton. S. 1–2. https://doi.org/10.1515/9783110307337.1.

Diewald, Gabriele (2018): Zur Diskussion: Geschlechtergerechte Sprache als Thema der germanistischen Linguistik – exemplarisch exerziert am Streit um das sogenannte generische Maskulinum. In: Zeitschrift für germanistische Linguistik 46(2), S. 283–299. https://doi.org/10.1515/zgl-2018-0016.

Diewald, Gabriele (2025a): Semantische Oppositionen und pragmatische Operationen: zur Bedeutungskonstitution und Verwendung von Personenbezeichnungen im Hinblick auf die semantische Domäne GESCHLECHT. In: Meuleneers, Paul/Zacharski, Lisa/Ferstl, Evelyn C./Nübling, Damaris (Hrg.): Genderbezogene Personenreferenzen: Routinen und Innovationen. Hamburg: Helmut Buske Verlag. S. 13–42. (= Linguistische Berichte, Sonderheft 36).

Diewald, Gabriele (2025b): Femininmovierung und movierte „Neografien": Überlegungen zu Oppositionstypen und Markiertheitswerten. In: Werth, Alexander (Hrg.): Die Movierung. De Gruyter. S. 13–42. https://doi.org/10.1515/9783111485102-002.

Eisenberg, Peter (2007): Sprachliches Wissen im Wörterbuch der Zweifelsfälle. Über die Rekonstruktion einer Gebrauchsnorm. In: Aptum 3, S. 209–228.

Eisenberg, Peter (2020): Zur Vermeidung sprachlicher Diskriminierung im Deutschen. In: Muttersprache 130(1), S. 3–16.

Eisenberg, Peter (2022): Weder geschlechtergerecht noch gendersensibel. In: APuZ Aus Politik und Zeitgeschichte 5–7, S. 30–35.

Garnham, Alan/Gabriel, Ute/Sarrasin, Oriane/Gygax, Pascal/Oakhill, Jane (2012): Gender Representation in Different Languages and Grammatical Marking on Pronouns: When Beauticians, Musicians, and Mechanics Remain Men. In: Discourse Processes 49(6), S. 481–500. https://doi.org/10.1080/0163853X.2012.688184.

Glim, Sarah/Körner, Anita/Härtl, Holden/Rummer, Ralf (2023): Early ERP Indices of Gender-Biased Processing Elicited by Generic Masculine Role Nouns and the Feminine–Masculine Pair Form. In: Brain and Language 242, S. 1–7. https://doi.org/10.1016/j.bandl.2023.105290.

Glück, Helmut (2020): Wissenschaftsfremder Übergriff auf die deutsche Sprache. Eine Kritik der „Handlungsempfehlungen der Bundeskonferenz der Frauen- und Gleichstellungsbeauftragten". In: Forschung & Lehre 12, S. 994–995.

Gygax, Pascal/Gabriel, Ute/Sarrasin, Oriane/Oakhill, Jane/Garnham, Alan (2008): Generically intended, but specifically interpreted: When beauticians, musicians, and mechanics are all men. In: Language and Cognitive Processes 23(3), S. 464–485. https://doi.org/10.1080/01690960701702035.

Gygax, Pascal/Garnham, Alan/Doehren, Sam (2016): What Do True Gender Ratios and Stereotype Norms Really Tell Us? In: Frontiers in Psychology 7, S. 1–6. https://doi.org/10.3389/fpsyg.2016.01036.

Haspelmath, Martin (2006): Against Markedness (and What to Replace It With). In: Journal of Linguistics 42(1), S. 25–70. https://doi.org/10.1017/S0022226705003683.

Haspelmath, Martin (2021): Explaining Grammatical Coding Asymmetries: Form–Frequency Correspondences and Predictability. In: Journal of Linguistics 57(3), S. 605–633. https://doi.org/10.1017/S0022226720000535.

Hellinger, Marlis/Bußmann, Hadumod (2003): Engendering Female Visibility in German. In: Hellinger, Marlis/Bußmann, Hadumod (Hrg.): Gender Across Languages: The Linguistic Representation of Women and Men. Volume 3. Amsterdam; Philadelphia: John Benjamins Publishing Co. S. 141–174.

Hetjens, Dominik/Hartmann, Stefan (2024): Effects of Gender Sensitive Language in Job Listings: A Study on Real-Life User Interaction. In: PLOS ONE 19(8), S. 1–18. https://doi.org/10.1371/journal.pone.0308072.

Klosa, Annette (2011): elexiko - ein Bedeutungswörterbuch zwischen Tradition und Fortschritt. In: Sprachwissenschaft 36(2/3), S. 275–306.

Kopf, Kristin (2023): Normalfall Movierung: Geschichte und Gegenwart des generischen Maskulinums in Prädikativkonstruktionen. In: Jahrbuch für Germanistische Sprachgeschichte 14(1), S. 196–226. https://doi.org/10.1515/jbgsg-2023-0014.

Körner, Anita (2025): Wertlose „Psychotests"? Psycholinguistische Experimente zu Geschlechtsassoziationen beim Lesen linguistischer Formen. In: Meuleneers, Paul W./Zacharski, Lisa/Ferstl, Evelyn C./Nübling, Damaris (Hrg.): Genderbezogene Personenreferenzen: Routinen und Innovationen. Helmut Buske Verlag. S. 177–196. (= Linguistische Berichte, Sonderheft 36).

Körner, Anita/Abraham, Bleen/Rummer, Ralf/Strack, Fritz (2022): Gender Representations Elicited by the Gender Star Form. In: Journal of Language and Social Psychology 41(5), S. 553–571. https://doi.org/10.1177/0261927X221080181.

Kotthoff, Helga/Nübling, Damaris (2024a): Genderlinguistik: Eine Einführung in Sprache, Gespräch und Geschlecht. 2. Auflage. Tübingen: Narr.

Kotthoff, Helga/Nübling, Damaris (2024b): Genderlinguistik: Eine Einführung in Sprache, Gespräch und Geschlecht. 2. Auflage. Tübingen: Narr Francke Attempto.

Krome, Sabine (2021): Gendern zwischen Sprachpolitik, orthografischer Norm, Sprach- und Schreibgebrauch. Bestandsaufnahme und orthografische Perspektiven zu einem umstrittenen Thema. In: Sprachreport 37(2), S. 22–29. https://doi.org/10.14618/sr-2-2021-krom.

Kupietz, Marc/Belica, Cyril/Keibel, Holger/Witt, Andreas (2010): The German Reference Corpus DeReKo: A primordial sample for linguistic research. Calzolari, Nicoletta/Tapias, Daniel/Rosner, Mike/Piperidis, Stelios/Odjik, Jan/Mariani, Joseph et al. (Hrg.): Proceedings of the Seventh conference on International Language Resources and Evaluation. International Conference on Language Resources and Evaluation (LREC-10). Valetta, Malta: European Language Resources Association (ELRA). S. 1848–1854.

Kupietz, Marc/Lüngen, Harald/Kamocki, Pawel/Witt, Andreas (2018): The German Reference Corpus DeReKo: New Developments – New Opportunities. Calzolari, Nicoletta/Choukri, Khalid/Cieri, Christopher/Declerck, Thierry/Goggi, Sara/Hasida, Koiti et al. (Hrg.): Proceedings of the Eleventh International Conference on Language Resources and Evaluation (LREC 2018). Miyazaki, Japan: European Language Resources Association (ELRA).

Link, Sabrina (2024): The Use of Gender-Fair Language in Austria, Germany, and Switzerland: A Contrastive, Corpus-Based Study. In: Lingua 308, S. 1–16. https://doi.org/10.1016/j.lingua.2024.103787.

Meineke, Eckhard (2023): Studien zum genderneutralen Maskulinum. Heidelberg: Universitätsverlag Winter.

Müller-Spitzer, Carolin (2022a): Der Kampf ums Gendern. Kontextualisierung der Debatte um eine geschlechtergerechte Sprache. In: Kursbuch 209(3), S. 28–45.

Müller-Spitzer, Carolin (2022b): Zum Stand der Forschung zu geschlechtergerechter Sprache. In: APuZ Aus Politik und Zeitgeschichte 5–7, S. 23–29.

Müller-Spitzer, Carolin/Ochs, Samira (2023): Geschlechtergerechte Sprache auf den Webseiten deutscher, österreichischer, schweizerischer und Südtiroler Städte. In: Sprachreport 39(2), S. 1–5. https://doi.org/10.14618/sr-2-2023_mue.

Müller-Spitzer, Carolin/Ochs, Samira/Koplenig, Alexander/Rüdiger, Jan Oliver/Wolfer, Sascha (2024): Less than One Percent of Words Would Be Affected by Gender-Inclusive Language in German Press Texts. In: Humanities and Social Sciences Communications 11(1), S. 1–13. https://doi.org/10.1057/s41599-024-03769-w.

Müller-Spitzer, Carolin/Ochs, Samira/Rüdiger, Jan Oliver/Wolfer, Sascha (2025): Die Herausbildung neuer Routinen der Personenreferenz am Beispiel der deutschen Weihnachts- und Neujahrsansprachen. In: Meuleneers, Paul W./Zacharski, Lisa/Ferstl, Evelyn C./Nübling, Damaris (Hrg.): Genderbezogene Personenreferenzen: Routinen und Innovationen. Hamburg: Helmut Buske Verlag. S. 213–236. (= Linguistische Berichte, Sonderheft 36).

Ochs, Samira (2025): *Ärztinnen und Pfleger, Biologen oder Chemikerinnen*: Alternierende Doppelformen als Mittel genderinklusiver Sprache in *Die Zeit*. In: Deutsche Sprache 2/2025, S. 97–125. https://doi.org/10.37307/j.1868-775X.2025.02.02.

Ochs, Samira/Rüdiger, Jan Oliver (2025): Of stars and colons: A corpus-based analysis of gender-inclusive orthographies in German press texts. In: Schmitz, Dominic/Stein, Simon David/Schneider, Viktoria (Hrg.): Linguistic intersections of language and gender. Düsseldorf: Düsseldorf University Press.

Pafel, Jürgen (2020): Referenz, Bd. 22. Heidelberg: Winter. (= Kurze Einführungen in die germanistische Linguistik).

Pettersson, Magnus (2011): Geschlechtsübergreifende Personenbezeichnungen: Eine Referenz- und Relevanzanalyse an Texten. Narr Francke Attempto Verlag.

Pusch, L. (1984): Das Deutsche als Männersprache. Frankfurt: edition suhrkamp.

Pustka, Elissa (2013): Quantifikator. Wörterbücher zur Sprach- und Kommunikationswissenschaft (WSK) Online. Berlin ; Boston: De Gruyter. https://www.degruyter.com/database/WSK/entry/wsk_idcc14eaad-bb56-4b35-b8ca-24edadcd2dee/html(28.11.2024).

Reifegerste, Jana/Meyer, Antje S./Zwitserlood, Pienie (2017): Inflectional complexity and experience affect plural processing in younger and older readers of Dutch and German. In: Language, Cognition and Neuroscience 32(4), S. 471–487. https://doi.org/10.1080/23273798.2016.1247213.

Reineke, Silke/Deppermann, Arnulf/Schmidt, Thomas (2023): Das Forschungs- und Lehrkorpus für Gesprochenes Deutsch (FOLK). In: Deppermann, Arnulf/Fandrych, Christian/Kupietz, Marc/Schmidt, Thomas (Hrg.): Korpora in der germanistischen Sprachwissenschaft. De Gruyter. S. 71–102. https://doi.org/10.1515/9783111085708-005.

Rothmund, Jutta/Scheele, Brigitte (2004): Personenbezeichnungsmodelle auf dem Prüfstand. In: Zeitschrift für Psychologie 212(1), S. 40–54. https://doi.org/10.1026/0044-3409.212.1.40.

Rüdiger, Jan Oliver (2025): QuickAnnotator. LINDAT/CLARIAH-CZ digital library at the Institute of Formal and Applied Linguistics (ÚFAL). http://hdl.handle.net/11234/1-5965(9.9.2025).

Schmitz, Dominic (2024): Instances of Bias: The Gendered Semantics of Generic Masculines in German Revealed by Instance Vectors. In: Zeitschrift für Sprachwissenschaft 43(2), S. 295–325. https://doi.org/10.1515/zfs-2024-2010.

Schmitz, Dominic/Schneider, Viktoria/Esser, Janina (2023): No Genericity in Sight: An Exploration of the Semantics of Masculine Generics in German. In: Glossa Psycholinguistics 2(1), S. 1–33. https://doi.org/10.5070/G6011192.

Simon, Horst J. (2022): Sprache Macht Emotion. In: APuZ Aus Politik und Zeitgeschichte 5–7, S. 16–22.

Sökefeld, Carla/Andresen, Melanie/Binnewitt, Johanna (2023): Personal noun detection for German. Proceedings of the 19th Joint ACL – ISO Workshop on Interoperable Semantic Annotation (ISA-19), Nancy, 20 June 2023. S. 33–39. https://sigsem.uvt.nl/isa19/ISA-19-proceedings.pdf.

Storjohann, Petra (2005): elexiko: A Corpus-Based Monolingual German Dictionary. In: Hermes, Journal of Linguistics 34, S. 55–82.

Trutkowski, Ewa/Weiß, Helmut (2023): Zeugen gesucht! Zur Geschichte des generischen Maskulinums im Deutschen. In: Linguistische Berichte 273, S. 5–39.

Waldendorf, Anica (2023): Words of change: The increase of gender-inclusive language in German media. In: European Sociological Review 40(2), S. 1–18. https://doi.org/10.1093/esr/jcad044.

Wegener, Heide (2024): Untersuchungen zur Interpretation generischer Maskulina - die Tests. In: Trutkowski, Ewa/Meinunger, André (Hrg.): Gendern - Auf Teufel*in komm raus? Berlin: Kulturverlag Kadmos. S. 33–57.

Zacharski, Lisa/Ferstl, Evelyn C. (2023): Gendered Representations of Person Referents Activated by the Nonbinary Gender Star in German: A Word-Picture Matching Task. In: Discourse Processes 60(4–5), S. 294–319. https://doi.org/10.1080/0163853X.2023.2199531.

Zifonun, Gisela (2018): Die demokratische Pflicht und das Sprachsystem: erneute Diskussion um einen geschlechtergerechten Sprachgebrauch. In: Sprachreport 34(4), S. 44–56.

*Alle URLs wurden am 08.04.2025 zum letzten Mal überprüft.*

**Adressen Autor*innen**

Carolin Müller-Spitzer/Samira Ochs/Jan Oliver Rüdiger/Sascha Wolfer

Institut für Deutsche Sprache

R 5, 6-13

D-68161 Mannheim

Corresponding Author: Carolin Müller-Spitzer; mueller-spitzer(at)ids-mannheim.de; https://orcid.org/0000-0002-5690-7774.